\documentclass{article} 
\usepackage{iclr2025_conference,times}

\usepackage{amsmath,amsfonts,bm}

\def\eqref#1{equation~\ref{#1}}

\def\1{\bm{1}}

\DeclareMathAlphabet{\mathsfit}{\encodingdefault}{\sfdefault}{m}{sl}
\SetMathAlphabet{\mathsfit}{bold}{\encodingdefault}{\sfdefault}{bx}{n}

\footnotetext[2]{This research is mainly done during their internship in SenseTime. }
\usepackage{hyperref}
\usepackage{url}

\usepackage{caption}
\usepackage{subcaption}
\usepackage{float}
\usepackage{placeins}
\usepackage{graphicx}
\usepackage{makecell}
\usepackage{stfloats}
\usepackage{booktabs}

\usepackage{multirow}
\usepackage{array} 
\usepackage[export]{adjustbox}
\usepackage{pifont}
\title{CVD-STORM: Cross-View Video Diffusion with Spatial-Temporal Reconstruction Model for Autonomous Driving}

\author{Tianrui Zhang$^{2\dagger}$\thanks{ Equal contribution.}  , Yichen Liu$^{1}$\footnotemark[1],   Zilin Guo$^{2\dagger}$\footnotemark[1],   Yuxin Guo$^{1}$, Jingcheng Ni$^{1}$, \\
\textbf{Chenjing Ding}$^{1}$, \textbf{Dan Xu}$^{2}$, \textbf{Lewei Lu}$^{1}$, \textbf{Zehuan Wu}$^{1}$  \\
\texttt{\{liuyichen,nijingcheng,guoyuxin,dingchenjing,luotto,wuzehuan\}@sensetime.com} \\
\texttt{\{tzhangbu,zguobd\}@connect.ust.hk}, \texttt{danxu@cse.ust.hk} \\
$^{1}$Sensetime Research, $^{2}$The Hong Kong University of Science and Technology \\
}

\def\ours{CVD-STORM}

\iclrfinalcopy 
\begin{document}

\maketitle

\begin{abstract}
Generative models have been widely applied to world modeling for environment simulation and future state prediction. With advancements in autonomous driving, there is a growing demand not only for high-fidelity video generation under various controls, but also for producing diverse and meaningful information such as depth estimation. To address this, we propose \ours, a cross-view video diffusion model utilizing a spatial-temporal reconstruction Variational Autoencoder (VAE) that generates long-term, multi-view videos with 4D reconstruction capabilities under various control inputs. Our approach first fine-tunes the VAE with an auxiliary 4D reconstruction task, enhancing its ability to encode 3D structures and temporal dynamics. Subsequently, we integrate this VAE into the video diffusion process to significantly improve generation quality. Experimental results demonstrate that our model achieves substantial improvements in both FID and FVD metrics. Additionally, the jointly-trained Gaussian Splatting Decoder effectively reconstructs dynamic scenes, providing valuable geometric information for comprehensive scene understanding. Our project page is \href{https://sensetime-fvg.github.io/CVD-STORM/}{https://sensetime-fvg.github.io/CVD-STORM}.
\end{abstract}
\begin{figure}[h]
    \centering
    \captionsetup[subfloat]{position=bottom, justification=justified, singlelinecheck=false}
    \captionsetup[subfloat]{skip=0.01cm} 
    \subfloat[Ground Truth]{\includegraphics[width = 1.01\linewidth]{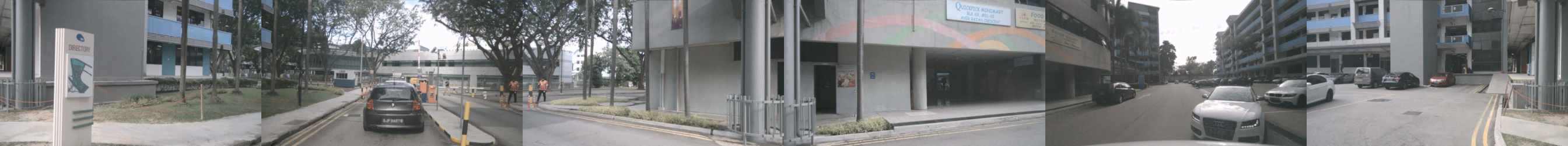}} 
    \vspace{0.01cm}
    \subfloat[w/o STORM-VAE (1200 step)]{\includegraphics[width = 1.01\linewidth]{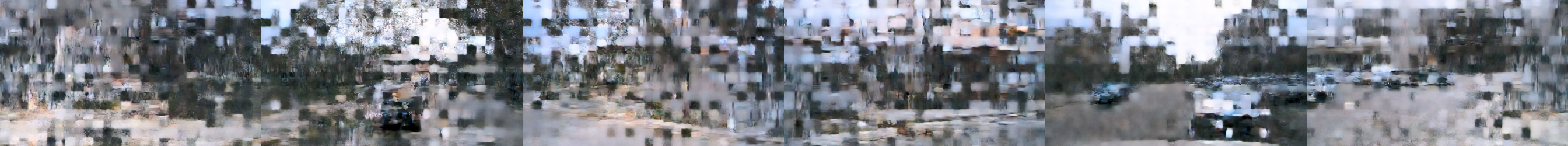}}
    \vspace{0.01cm}
    \subfloat[w/ STORM-VAE (1200 step)]{\includegraphics[width = 1.01\linewidth]{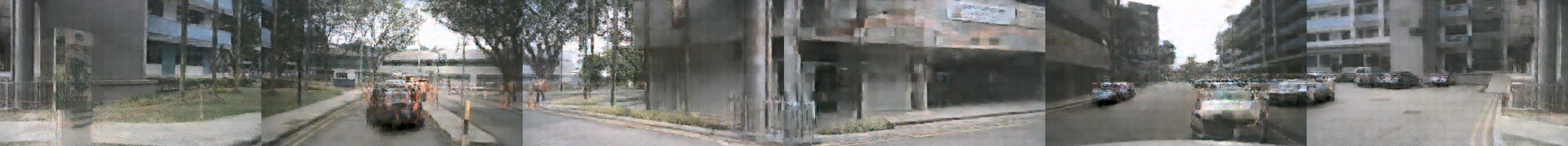}} 
    \vspace{-0.2cm}
    \caption{\textbf{Early-Stage Generation Visualization.}
    (a) shows the ground-truth sequence.
    (b) depicts the model’s output at training step 1,200 when using a standard VAE.
    (c) presents the corresponding output generated with our STORM-VAE at the same step.
    Notably, (c) exhibits significantly improved convergence and visual fidelity compared to (b), demonstrating the effectiveness of our approach even at early stage in training.}
    \label{fig:teaser}
\end{figure}

\section{Introduction}
\label{sec:intro}
Autonomous vehicles have emerged as a prominent research domain within artificial intelligence applications. The development of reliable self-driving systems necessitates both extensive data collection for training decision-making algorithms and sophisticated closed-loop simulations to verify planning outputs. These requirements present significant challenges, particularly the need for driving world models that accurately represent the environment and enables precise prediction of future scenarios. Concurrently, diffusion models have become the state-of-the-art approach for video generation, offering a promising solution for realistic simulation. Recent advances in this field have demonstrated the remarkable capability of these models to generate photorealistic videos~\cite{cogvideo,opensora, opensora2}, with successful applications extending to complex driving scenarios~\cite{drivegan,drivedreamer2}. 

To serve as comprehensive driving world models, diffusion-based approaches must be capable of generating long-term, multi-view, and controllable videos. Early attempts such as \cite{magicdrive, drivegan} struggled with generating extended sequences and following complex conditional inputs. Recent advancements, however, have significantly addressed these limitations by adopting architectures and methodologies from high-performing diffusion models.
For example, \cite{unimlvg, magicdrivev2, cosmos-drive-dreams} have all implemented spatial-temporal diffusion transformer (DiT) architectures and employed multi-stage training strategies, progressively enhancing generative fidelity and temporal consistency. Nevertheless, despite incorporating cross-view generation, these approaches lack explicit 3D information, which constrains their applicability as world models.
To overcome this limitation, \cite{gao2024magicdrive3d} directly applies an enhanced 3D Gaussian Splatting (3DGS) technique to diffusion outputs, though internal inconsistencies in the generated images remain inadequately resolved. UniScene \cite{li2025uniscene} incorporates semantic occupancy as conditional guidance for LiDAR generation, but requires additional annotation during the training process. Other approaches \cite{hassan2025gem, unifuture} produce depth maps supervised by Depth Anything V2 \cite{yang2024depth}, but these relative depth estimates cannot accurately represent real-world geometry. While \cite{holodrive} attempts to generate LiDAR data and video simultaneously, the LiDAR is not well aligned with the images.

To address these challenges, we propose \ours, a framework that generates long sequential multi-view driving videos while simultaneously decoding reconstructed scenes represented by dynamic 3D Gaussian Splatting (3DGS) \cite{3dgs}. First, we finetune an image VAE with an affiliated Gaussian decoder as described in STORM \cite{yang2025storm}, enabling the decoding of VAE latents into 3D Gaussians.
This finetuned model, dubbed as STORM-VAE, serves as the latent encoder for training a cross-view video diffusion model with the same architecture as \cite{unimlvg}.
Recent research \cite{repa, repae, fuest2024diffusion} has established that representation learning is crucial to diffusion model performance. Aligned with these findings, our experiments demonstrate that the latents encoded by STORM-VAE, which fuse information from LiDAR and across frames, significantly improve the generative quality and convergence rate. Figure \ref{fig:teaser} illustrates the impressive denoising ability of \ours at an early step, compared with the one without STORM-VAE.
During inference, \ours~can generate long-term six-view videos conditioned on text, bounding boxes (BBox), and high-definition maps (HDMap), while the Gaussian Splatting (GS) Decoder can directly reconstruct 4D scenes from the generated latents.

In summary, our main contributions are:
\begin{itemize}
\item We introduce STORM-VAE, an extended VAE model incorporating a Gaussian Splatting decoder for 4D scene reconstruction. This auxiliary network integrates spatial and temporal information into the latent representation, moving beyond RGB-only encoding. Meanwhile, it can also achieve 4D reconstruction in the driving scenarios.
\item We propose \ours, a novel pipeline for driving world modeling that simultaneously generates multi-view videos and reconstructs 4D scenes. To manage the complexity of these tasks, we adopt a two-stage training strategy: first learning scene reconstruction, followed by training a conditional world model.
\item Our experiments demonstrate that \ours ~not only significantly improve the generative quality of the current world model by enhancing representation learning,  but also addresses the challenges of 4D absolute depth estimation.
\end{itemize}
\section{Related Work}
\label{sec:related}

\subsection{Video Diffusion and Driving World Model}
The diffusion approach has become the mainstream for generative tasks. With the advancements in 2D image diffusion models~\cite{stable-diffusion, controlnet, flux1, hunyuan-dit}, this technique has rapidly extended to video generation~\cite{cogvideo, cogvideox, seedance, opensora, opensora2, hunyuanvideo}, yielding impressive visualizations and enabling precise control. In addition, related studies highlight its significant potential as a real-world simulator.

In the field of autonomous driving, research started to focus on constructing driving world models based on video generation to simulate realistic driving scenarios. For instance, GenAD~\cite{genad} leverages large-scale web video datasets to enhance long-duration video generation capabilities, while Vista~\cite{vista} incorporates action inputs to control vehicle trajectories. However, these approaches are limited to single-view generation and do not include other conditions to simulate road conditions. There still exists a significant gap between their capabilities and real-world driving requirements.

Therefore, generating multi-view video with precise control and long-term consistency has attracted significant research attention. Early approaches such as~\cite{magicdrive, drivedreamer2, xie2025glad} achieved promising results for short-term videos but struggled to extend sequence length effectively. The emergenece of DiT~\cite{dit} substantially improved diffusion model scalability, prompting numerous researchers to incorporate transformer architectures into driving world models. UniMLVG~\cite{unimlvg} enhancs Stable Diffusion 3.5 \cite{sd35} with temporal and multi-view modules, successfully unifying multiple datasets with heterogeneous structures during training. Similarly, MagicDriveV2 \cite{magicdrivev2} also employs this design but encodes videos through 3D VAE to achieve greater data compression. This architecture has demonstrated exceptional performance when applied to larger-scale datasets \cite{cosmos-drive-dreams, gaia2}. Additionally, researchers also have successfully incorporated action control mechanisms to enable the generation of precisely controllable multi-view videos~\cite{maskgwm}. Despite these advancements, current generative methods still fail to adequately capture important structural information, particularly depth data.

\subsection{4D Reconstruction in Driving Scenarios}
Capturing 3D information is crucially important in driving scenarios and numerous studies has explored how to predict the depth or reconstruct the 4D scene in the front-view driving videos. Some research incorporates the structure prediction in the generative procedure. For instance,  UniFuture~\cite{unifuture} directly unified the depth prediction into the video generation to attain highly aligned RGB-Depth correspondence. However, this work needs Depth Anything V2~\cite{yang2024depth} to generate pseudo supervision for depth. Additionally, this approach can only produce relative depth, which is insufficient for the autonomous driving application. Another unified framework GEM~\cite{hassan2025gem} mitigates problems with consistencies in long-range video generation, yet still preserves similar problem in depth estimation as UniFuture. \\

On the other hand, a considerable body of research has focused on incorporating reconstruction tasks into driving scenarios. MagicDrive3D~\cite{gao2024magicdrive3d} employs a two-stage pipeline that integrates Gaussian splatting for 3D reconstruction. Although presented as a unified framework, the second-stage reconstruction process exerts minimal influence on the generative model in the first stage, limiting true end-to-end interaction. More approaches concentrate primarily on pure reconstruction objectives. For instance, ReconDreamer~\cite{ni2025recondreamer} introduces a network trained to correct artifacts in novel views reconstructed from a pretrained 3D Gaussian representation. Similarly, OmniScene~\cite{wei2025omni} leverages forward Gaussian mapping to obtain a 3D scene representation in bird's-eye view (BEV) format. Building upon this, STORM~\cite{yang2025storm} advances the paradigm by employing forward 4D Gaussian splatting to capture spatiotemporal dynamics through sequential scene modeling.

\begin{figure*}[t]
\centering
    \includegraphics[width=0.99\linewidth]{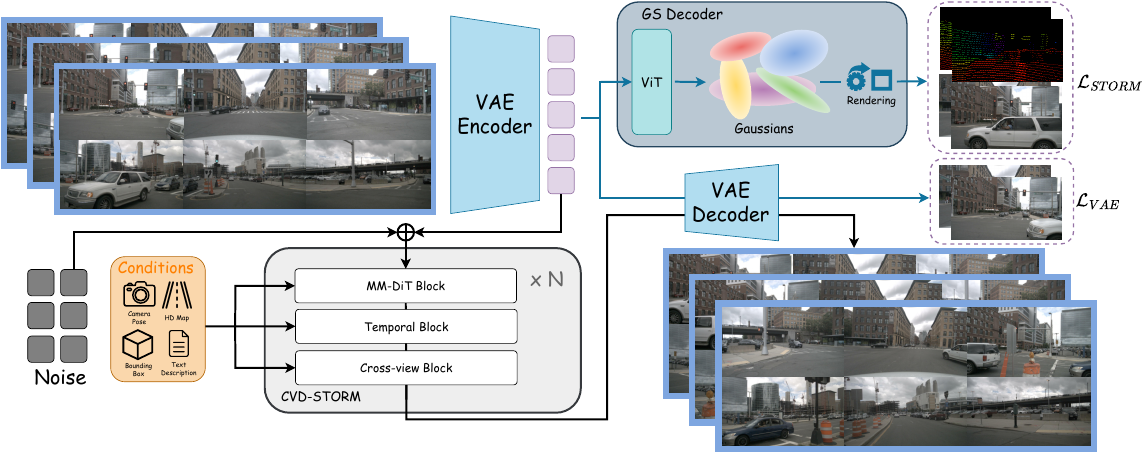}
    \caption{\textbf{Overall framework of the model. } Our pipeline contains two models. The upper section illustrates STORM-VAE training, with the forward process indicated by blue arrows. STORM-VAE takes multi-view images from context timesteps and processes the image latents through two decoders: the VAE Decoder performs image reconstruction (updated by $\mathcal L_{\text{VAE}}$), while the GS Decoder performs scene reconstruction (updated by $\mathcal L_{\text{STORM}}$). The lower section illustrates the inference pipeline of \ours , with the forward process shown by solid block arrows. The diffusion part can either use STORM-VAE latents as reference frames for prediction or generate from noise, while incorporating various conditioning inputs for guidance.
    }
    \vspace{-4mm}
    \label{fig:pipeline}
\end{figure*}

While both generative modeling and 3D reconstruction have been extensively studied in autonomous driving contexts, few works have explored the integration of these two tasks in a synergistic manner. The potential of jointly optimizing generation and reconstruction remains largely underexplored.

\subsection{Representation Learning in Diffusion}
Recent research has devoted considerable effort to exploring better latent representations for improving diffusion model performance \cite{fuest2024diffusion}. \cite{yang2022your, addp, deja2023learning} involves incorporating additional tasks during generation training, such as classification and segmentation. Other works focuses on aligning the latent space with that of foundation models. For example, \cite{pernias2023wurstchen} divides diffusion training into two stages, with the first stage dedicated to training an additional encoder that extracts image semantic features. REPA \cite{repa} takes intermediate features from the diffusion model and projects them to align with features from pretrained models, while VA-VAE \cite{vavae} performs this alignment during variational auto-encoder (VAE) training. Building upon REPA, REPA-E \cite{repae} finetunes the entire model end-to-end, allowing the alignment loss to update VAE parameters and thereby accelerating generation performance.

Inspired by these advances, we extend this representation learning approach to video diffusion models by introducing a reconstruction task during training and simultaneously tuning the VAE. This approach aims to enhance generation performance while achieving a significant additional capability --- 4D reconstruction.

\section{Method} 
\label{sec:method}

Figure \ref{fig:pipeline} illustrates the overall pipeline of our proposed method. Our framework generates multi-view driving videos conditioned on various inputs, including text prompts, bounding boxes, HD maps—with or reference frames, while simultaneously producing scene reconstructions represented as dynamic 3DGS.
Our approach extends UniMLVG \cite{unimlvg} by enhancing its variational autoencoder (VAE) architecture and refining the training procedure. Specifically, we first finetune the pretrained image VAE to create STORM-VAE, which incorporates an additional reconstruction task adapted from STORM \cite{yang2025storm}. This modification introduces a Gaussian Decoder capable of reconstructing 3D Gaussians and their associated velocities. We then leverage STORM-VAE to train a DiT-based diffusion model that employs three distinct transformer blocks operating along different data dimensions, which improve both spatial coherence and temporal consistency in the generated outputs.
\subsection{Preliminary: STORM} 
\label{sec:method:storm}
Given a set of images $\{\bm{I}_t^v \in \mathbb{R}^{H\times W\times 3}\}$ with corresponding camera poses from timestamps $t \in T_c$ and viewpoints $v \in V$, STORM fuses image features through a Vision Transformer (ViT) and generate pixel-level Gaussians $\{\bm{G}_t^v \in \mathbb{R}^{H\times W \times 12}\}$. Each Gaussian is characterized by its center $\bm{\mu} \in \mathbb{R}^3$, orientation $\mathcal{R} \in \mathbb{SO}(3)$, scale $s \in \mathbb{R}$, opacity $o \in \mathbb{R}$, and color $\bm{c} \in \mathbb{R}^3$. The center $\bm{\mu}$ is positioned along the ray cast from the camera center, allowing the Gaussian decoder to only output the depth value. Additionally, the model predicts the velocity of each Gaussian to model dynamic scene elements. To render target viewpoints at timestamp $t'$, the 3D Gaussian Splatting (3DGS) elements $\bm{G}_t^v$ are transformed according to their predicted velocities into Gaussians at time $t'$, denoted as $\bm{G}_{t \rightarrow t'}^v$. The target images are then rendered based on the union of all $\bm{G}_{t\rightarrow t'}^v$. To enhance image quality, STORM incorporates auxiliary tokens to compose sky colors and adopts view-based exposure variations.

The training process is supervised by target views randomly sampled within a predefined sampling range. The image rendering loss $\mathcal{L}_{\text{rgb}}$ is formulated as:
\begin{equation}
    \mathcal{L}_{\text{rgb}} = \sum_{t'\in T_t, v \in V} \| D(F(\{\bm{I}_t^v\}), t', v) - \bm{I}_{t'}^v \|_2^2,
\end{equation}
where $F$ represents the ViT encoder, $D$ denotes the decoder and rendering, including all image post-processing operations, and $T_t$ is the set of target timesteps. Additionally, the Gaussian rendering can also produce depth so we use the depth map obtained by projecting LiDAR on camera views to supervise the training. We define the overall loss as $\mathcal{L}_{\text{STORM}}$ and omit discussion of additional loss terms not directly relevant to this paper. For more detailed description of the methodology, please refer to \citet{yang2025storm}.
\subsection{STORM-VAE}
\label{sec:method:storm-vae}
We introduce STORM-VAE, a novel variational autoencoder that incorporates STORM as an auxiliary network within the VAE framework. The upper part of Figure \ref{fig:pipeline} illustrates the architecture of our proposed model. STORM-VAE builds upon a general VAE structure, specifically utilizing the pretrained VAE from Stable Diffusion 3.5 (SD3.5) \cite{sd35} in our setting.
In the STORM-VAE pipeline, the VAE encoder $E$ first encodes input images into latent representations, which are subsequently processed through two parallel branches. In the first branch, the latents are processed by the VAE decoder $D_{VAE}$ to ensure high-fidelity image reconstruction, supervised by the loss function $\mathcal{L}_{\text{VAE}}$. In the second branch, sampled context latents are fed into the Gaussian Splatting decoder ($D_{\text{GS}}$), which shares architectural similarities with STORM. The key distinction is that STORM processes RGB images directly while the $D_{\text{GS}}$ operates on the VAE latent representations. Consequently, the new RGB rendering loss is formulated as:
\begin{equation}
\mathcal{L}_{\text{rgb}} = \sum_{t'\in T_t, v \in V} D_{\text{GS}}(E({\bm{I}t^v}), t', v) - \bm{I}{t'}v |_22,
\end{equation}
where $D_{\text{GS}}$ is equivalent to $D\cdot F$ described in Section \ref{sec:method:storm}.
The comprehensive training objective combines the VAE and STORM components as follows:
\begin{equation}
\mathcal{L} = \mathcal{L}_{\text{VAE}} + \lambda \mathcal{L}_{\text{STORM}}.
\end{equation}
Here, $\mathcal{L}_{\text{VAE}}$ comprises three components: the reconstruction loss $\mathcal{L}_{\text{MSE}}$, the perceptual loss $\mathcal{L}_{\text{LPIPS}}$, and the KL divergence loss $\mathcal{L}_{\text{KL}}$. We deliberately excluded the GAN loss from our implementation as our experiments indicated it compromised training stability. In our experiments, we set $\lambda$ to 0.5.

\subsection{\ours}
\label{sec:method:cvd}
The lower part of Figure \ref{fig:pipeline} illustrates the architecture of \ours. Following UniMLVG \cite{unimlvg}, we adopt SD3.5 as initialization and append a temporal block and a cross-view block after each Multi-Modality DiT (MM-DiT) block of SD3.5. The input latent of \ours~is $z_t \in \mathbb R^{T\times V\times C\times H\times W}$, where $T$ is the number of frames, $V$ is the number of viewpoints, $C$ is the latent dimension, and $H, W$ are the latent spatial dimensions of a single image. The MM-DiT block performs attention only at the image level (i.e., across $H \times W$ dimensions), which requires reshaping the input to $HW \times TV \times C$ before processing. Similarly, the temporal block operates on the sequence length dimension and the cross-view block operates on the view dimension, requiring to  reshape the input to $T \times VHW \times C$ and $V \times THW \times C$, respectively. We also incorporate the multiple conditioning approaches and multi-task framework from UniMLVG in our training. For details regarding these components, please refer to their paper. The training loss utilizes rectified flow \cite{liu2022flow}, formulated as:
\begin{equation}
\mathcal{L}_{\text{SD}} = \mathbb{E}_{\epsilon \sim \mathcal{N}(0, I)} \left[ \left\| \epsilon_{\theta}(z_t, t, c) - (z_0 - \epsilon) \right\|^2 \right],
\end{equation}
where $\epsilon_\theta$ denotes the model, $z$ represents the STORM-VAE latent, $z_t$ is the noisy latent, $\epsilon$ is the noise, $t$ is the timestep, and $c$ is the conditioning information.

Different from UniMLVG, we replace the SD3.5 VAE with our STORM-VAE, which provides enhanced latent representations and the capability to estimate absolute depth. Furthermore, rather than employing a multi-stage training process to progressively develop temporal and spatial generation capabilities, we jointly train the temporal blocks, spatial blocks, and MM-DiT blocks in a single stage. This integrated approach significantly simplifies the training procedure and reduces computational costs.
\begin{table*}[t]
\centering
\begin{tabular}{l|cccccc}
\toprule
\textbf{Method} &  \textbf{Duration} & \textbf{FID}$\downarrow$ & \textbf{FVD}$\downarrow$ & $\textbf{mAP}_{obj}$$\uparrow$ & $\textbf{mIoU}_{r}$$\uparrow$ & $\textbf{mIoU}_{v}$$\uparrow$\\ 
\midrule
DreamForge~\cite{dreamforge} &20s&16&224.8&13.80&-&-\\
UniScene~\cite{li2025uniscene}& - & 6.1 &70.5&-&-&-\\
Glad~\cite{xie2025glad}& - & 11.2&118.0&-&-&-\\
DriveScape~\cite{drivescape} &-&8.3&76.4& -& 64.43& 28.86\\
MagicDrive2~\cite{magicdrivev2}  & 5s & 19.1 & 218.1 & 12.30 & 61.05 & 27.01\\ 
DriveSphere~\cite{yan2025drivingsphere}  &-&-&103.4&21.45&-&-\\
DiVE~\cite{jiang2024dive}  &20s &-&94.6 & \underline{24.55}&-&- \\
UniMLVG~\cite{unimlvg} & 20s & 5.8 & 36.1 & 22.50 & \textbf{70.81} & \underline{29.12}  \\
\midrule

\textbf{\ours}  & 20s & \textbf{3.8} & \textbf{14.0} & \textbf{25.21} & \underline{66.11} & \textbf{29.84}  \\ 
\bottomrule
\end{tabular}
\caption{\textbf{Comparison of the generation quality and condition-following metrics on nuScenes validation set.} The best results are in \textbf{bold}, while the second best results are in \underline{underlined}. Since most of methods do not release their checkpoints, we list the results reported in their paper. $-$ represents the values not mentioned in the corresponding papers. $\text{mIoU}_{r}$ and $\text{mIoU}_{v}$ are the short of the mean IoU of road and vehicle. }
\vspace{-0.4cm}
\label{tab:main}
\end{table*}

\newcommand\rgbHeight{0.05cm}
\begin{figure}[t]
    \centering 
    \subfloat{\includegraphics[width = 0.166\linewidth]{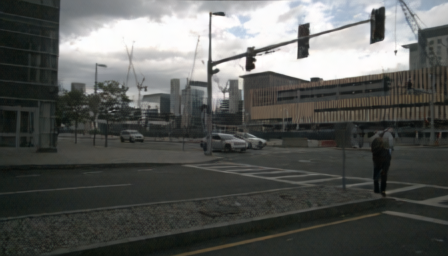}} 
    \subfloat{\includegraphics[width = 0.166\linewidth]{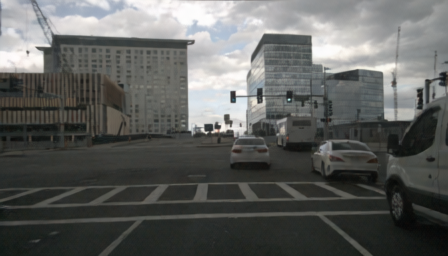}} 
    \subfloat{\includegraphics[width = 0.166\linewidth]{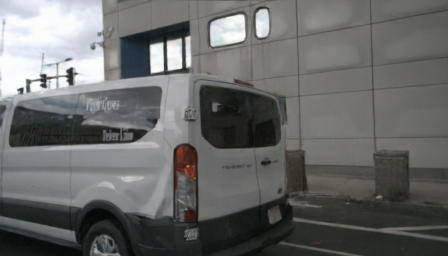}} 
    \subfloat{\includegraphics[width = 0.166\linewidth]{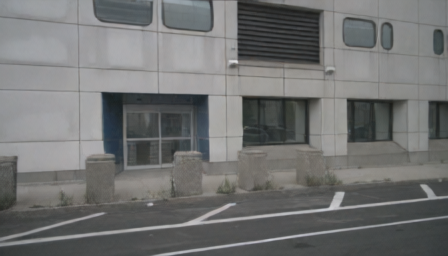}} 
    \subfloat{\includegraphics[width = 0.166\linewidth]{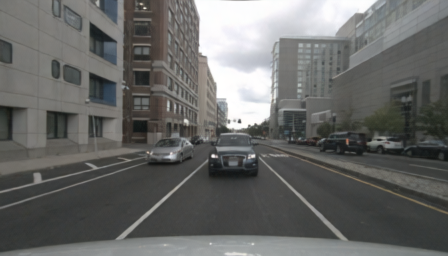}} 
    \subfloat{\includegraphics[width = 0.166\linewidth]{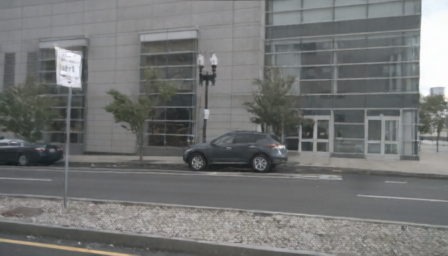}}
    \vspace{-0.05cm}
    \subfloat{\includegraphics[width = 0.166\linewidth]{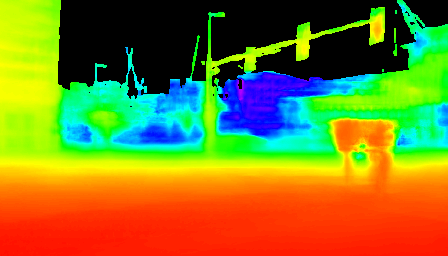}} 
    \subfloat{\includegraphics[width = 0.166\linewidth]{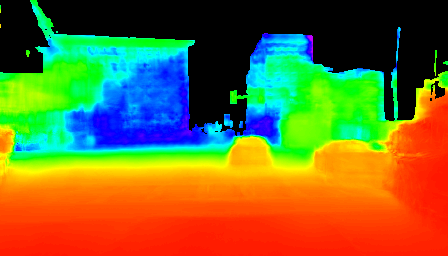}} 
    \subfloat{\includegraphics[width = 0.166\linewidth]{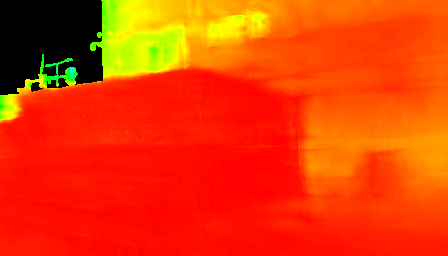}} 
    \subfloat{\includegraphics[width = 0.166\linewidth]{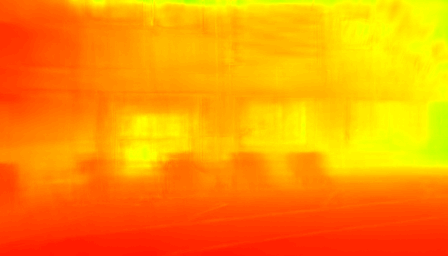}} 
    \subfloat{\includegraphics[width = 0.166\linewidth]{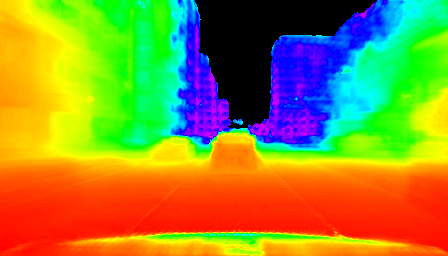}} 
    \subfloat{\includegraphics[width = 0.166\linewidth]{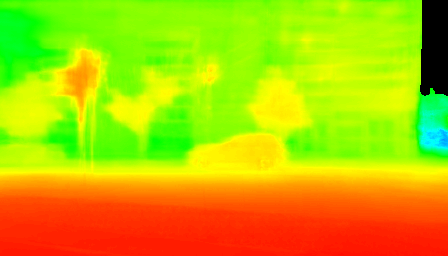}}    
    \vspace{0.1cm} 
    
    \subfloat{\includegraphics[width = 0.166\linewidth]{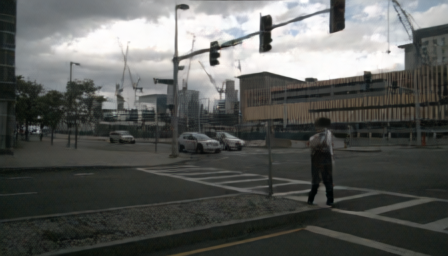}}
    \subfloat{\includegraphics[width = 0.166\linewidth]{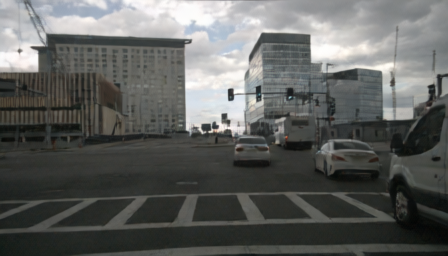}} 
    \subfloat{\includegraphics[width = 0.166\linewidth]{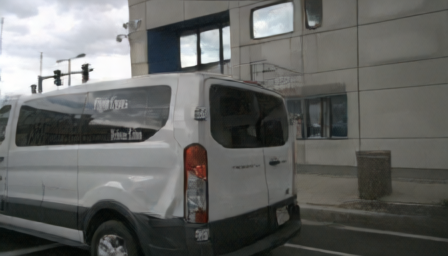}} 
    \subfloat{\includegraphics[width = 0.166\linewidth]{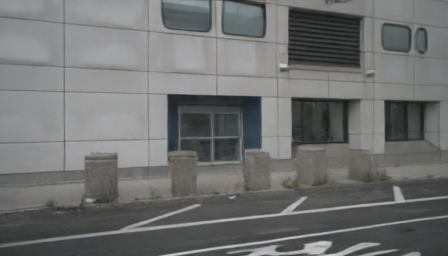}} 
    \subfloat{\includegraphics[width = 0.166\linewidth]{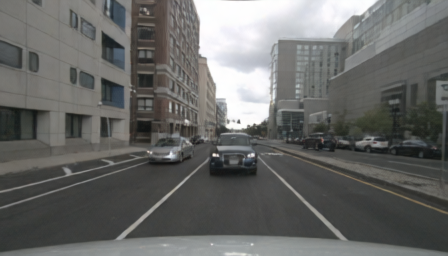}} 
    \subfloat{\includegraphics[width = 0.166\linewidth]{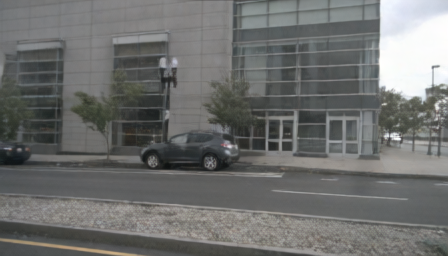}}
    \vspace{-0.05cm}
    \subfloat{\includegraphics[width = 0.166\linewidth]
    {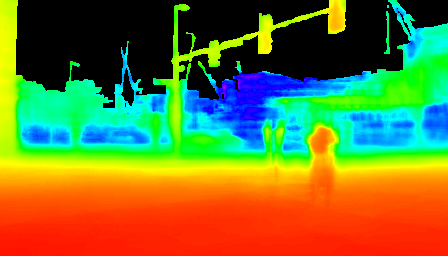}}
    \subfloat{\includegraphics[width = 0.166\linewidth]{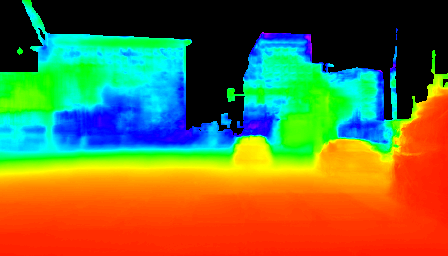}}
    \subfloat{\includegraphics[width = 0.166\linewidth]{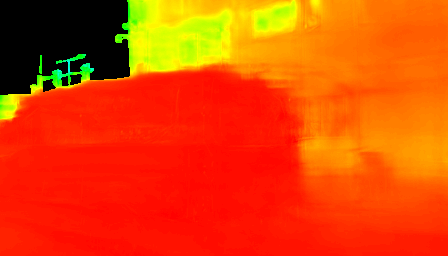}}
    \subfloat{\includegraphics[width = 0.166\linewidth]{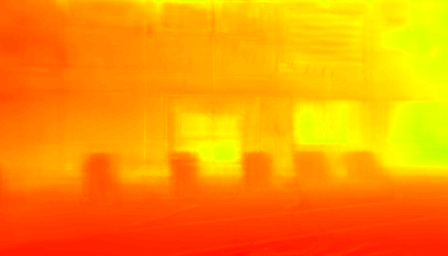}}
    \subfloat{\includegraphics[width = 0.166\linewidth]{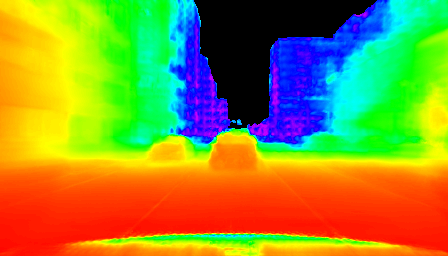}}
    \subfloat{\includegraphics[width = 0.166\linewidth]{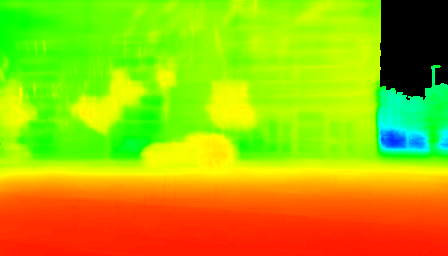}}    
    \vspace{0.1cm}

    \subfloat{\includegraphics[width = 0.166\linewidth]{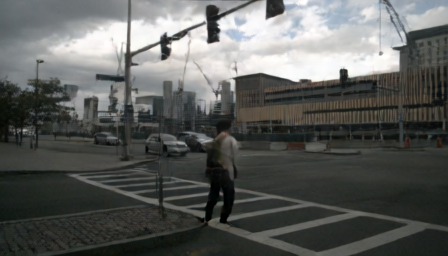}} 
    \subfloat{\includegraphics[width = 0.166\linewidth]{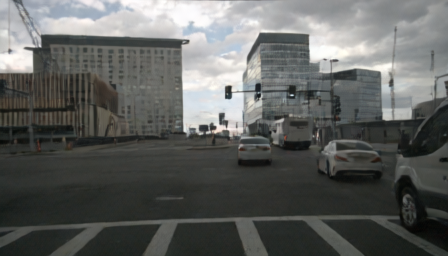}} 
    \subfloat{\includegraphics[width = 0.166\linewidth]{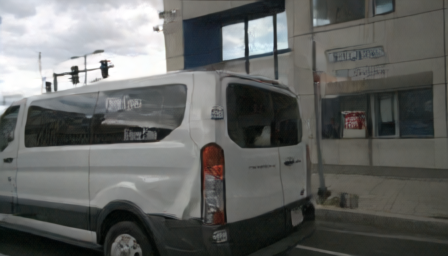}} 
    \subfloat{\includegraphics[width = 0.166\linewidth]{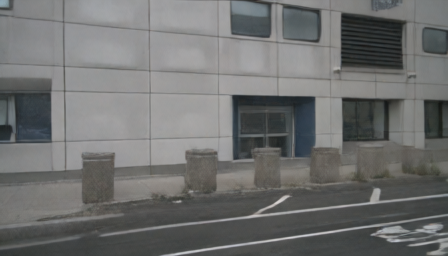}} 
    \subfloat{\includegraphics[width = 0.166\linewidth]{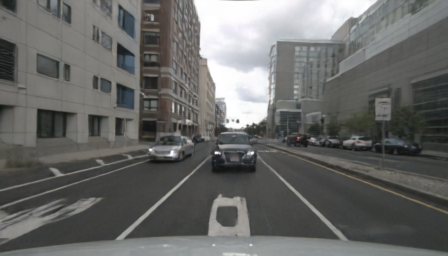}} 
    \subfloat{\includegraphics[width = 0.166\linewidth]{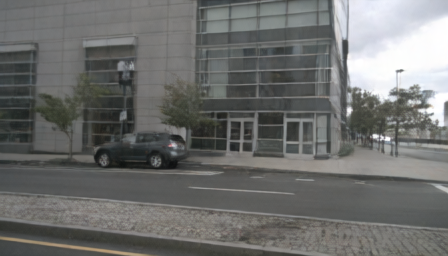}}
    \vspace{-0.05cm}
    \subfloat{\includegraphics[width = 0.166\linewidth]
    {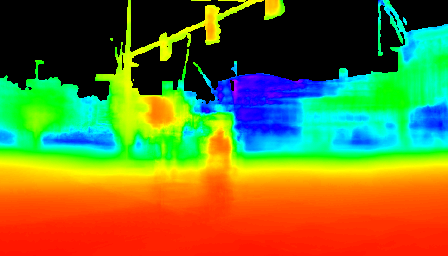}}
    \subfloat{\includegraphics[width = 0.166\linewidth]{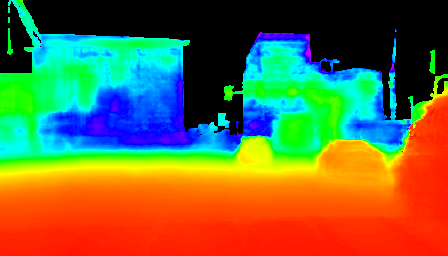}}
    \subfloat{\includegraphics[width = 0.166\linewidth]{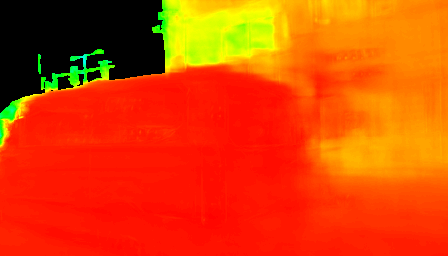}}
    \subfloat{\includegraphics[width = 0.166\linewidth]{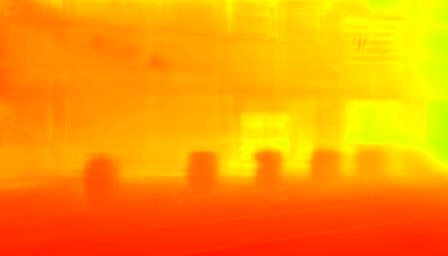}}
    \subfloat{\includegraphics[width = 0.166\linewidth]{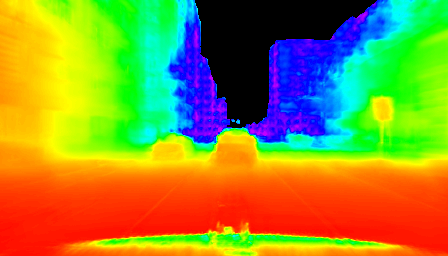}}
    \subfloat{\includegraphics[width = 0.166\linewidth]{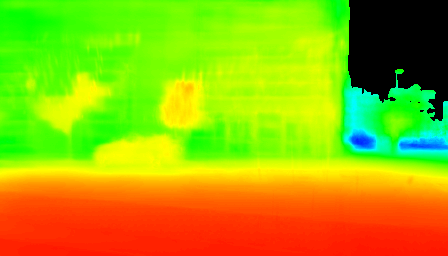}} 
    
    \vspace{-0.2cm}
    \caption{\textbf{Qualitative results of Depth Estimation. }This figure illustrates the depth of the videos generated by \ours  \;at frame 0, 5, 10. Our GS decoder can successfully extract the depth information of dynamic and static objects.}
    \label{fig:vis_depth}
\end{figure}

\section{Experiments}
\subsection{Experiment Details}
\subsubsection{Datasets}
We adopt both single-view and multi-view datasets in our training: OpenDV-Youtube \cite{yang2024generalized} for single-view data, and nuScenes \cite{caesar2020nuscenes}, Waymo \cite{sun2020scalability}, and Argoverse2 \cite{wilson2023argoverse} for multi-view data. We set the sequence length of a single simple as 19. To enhance the extensibility and diversity of our model, we incorporate three different image resolutions: $144 \times 256$, $176 \times 304$, and $256 \times 448$, with sampling ratios of 0.1, 0.3, and 0.6, respectively. All the models are trained on H100 with batch size 32. For diffusion training, we leverage available dataset annotations, including 3D bounding boxes, HD maps, and camera parameters. For nuScenes specifically, we utilize 12 Hz interpolated annotations. Text descriptions for all frames and views are generated at 2 Hz (key frames for evaluation). 
\subsubsection{Evaluation Metrics}
To assess the effectiveness of our method in terms of realism, continuity, ad precise control, we selected four key metrics to compare against existing multi-view image and video generation methods.
We use the widely recognized Fr\'echet Inception Distance(FID)~\cite{heusel2017gans} for realism evalution and Fr\'echet Video Distance (FVD)~\cite{unterthiner2018towards} for  temporal coherence estimation. 
To evaluate controllablity, we evaluate two perception tasks: 3D object detection and BEV segmentation of road maps. These tasks serve as proxies for measuring the spatial accuracy and consistent geometry of our generated content. We adopt BEVFormer~\cite{li2022bevformer} and  cross-view transformers~\cite{zhou2022cross} to evaluate the performance on these two tasks respectively.

\subsubsection{Implementation details}
For STORM-VAE training, we designate the 1st, 7th, 13th and 19th frames as the context frames while 3 timesteps are randomly sampled as targets. Since the Opendv-Youtube is a single-view dataset without LiDAR data, it is used exclusively to train the VAE image reconstruction branch. The other three datasets are utilized for both VAE and STORM training. To address viewpoint inconsistency across datasets, we standardize inputs to 6 views for all datasets and implement attention masking to avoid redundant data fusion.

For diffusion training, we freeze the encoder of STORM-VAE. As discussed in Section \ref{sec:method:cvd}, we implement the single-stage training so we have to deal with the invariance across datasets. For OpenDV-Youtube, the cross-view block is omitted due to its single-view nature. For multi-view datasets, we randomly drop temporal and cross-view blocks to enhance the generative capability of each individual block, thereby improving the overall model stability and robustness. During inference, we use 3 frames as reference for autoregressive prediction. A cosine scheduler is used with initial learning rate of $6 \times 10^{-5}$. The minimum learning rate is set to $1 \times 10^{-7}$. The optimize is widely used AdamW. The inference steps are set to 50. All Experiments are conducted on H100 GPUs.
\begin{table}[t]
    \caption{\textbf{Ablation Study }}
    \vspace{-0.2cm}
    \begin{subtable}[t]{.5\linewidth}
        \centering
        \begin{tabular}{c|cc}
        \toprule
        \textbf{\# Ref. frames} & FID & FVD \\ 
        \midrule
        0 & 8.7 & 39.0 \\
        1& \textbf{3.6} & 17.2 \\
        3  &  3.8 & \textbf{14.0} \\
        \bottomrule
        \end{tabular}
        \vspace{0.2in}
        \caption{\textbf{Ablation study of the number of reference frames}. The best results are in \textbf{bold}. The FID is about the same with reference frame, while FVD strictly decreases with larger reference frame count.}
        \vspace{-0.2in}
        \label{tab:ref_frame}
    \end{subtable}%
    \hspace{0.2cm}
    \begin{subtable}[t]{.5\linewidth}
        \centering
        \begin{tabular}{l|cc}
        \toprule
        \textbf{VAE used} & FID & FVD \\ 
        \midrule
        w/o STORM-VAE & 9.36 & 52.85 \\
        w/  STORM-VAE  &  \textbf{7.92} & \textbf{34.37} \\
        \bottomrule
        \end{tabular}
        \vspace{0.2in}
        \caption{\textbf{Ablation study of the use of VAE}. The best results are in \textbf{bold}. w/o STORM-VAE means using default vae of SD3.5. Both models are trained for 40k steps with Opendv, nuScenes, Waymo, Argoverse2. No pretarined weight is loaded for diffusion for fair comparison.}
        \vspace{-0.2in}
        \label{tab:storm_vae}
    \end{subtable}
\end{table}
\begin{figure}[t]
\centering
    \includegraphics[width=0.99\linewidth]{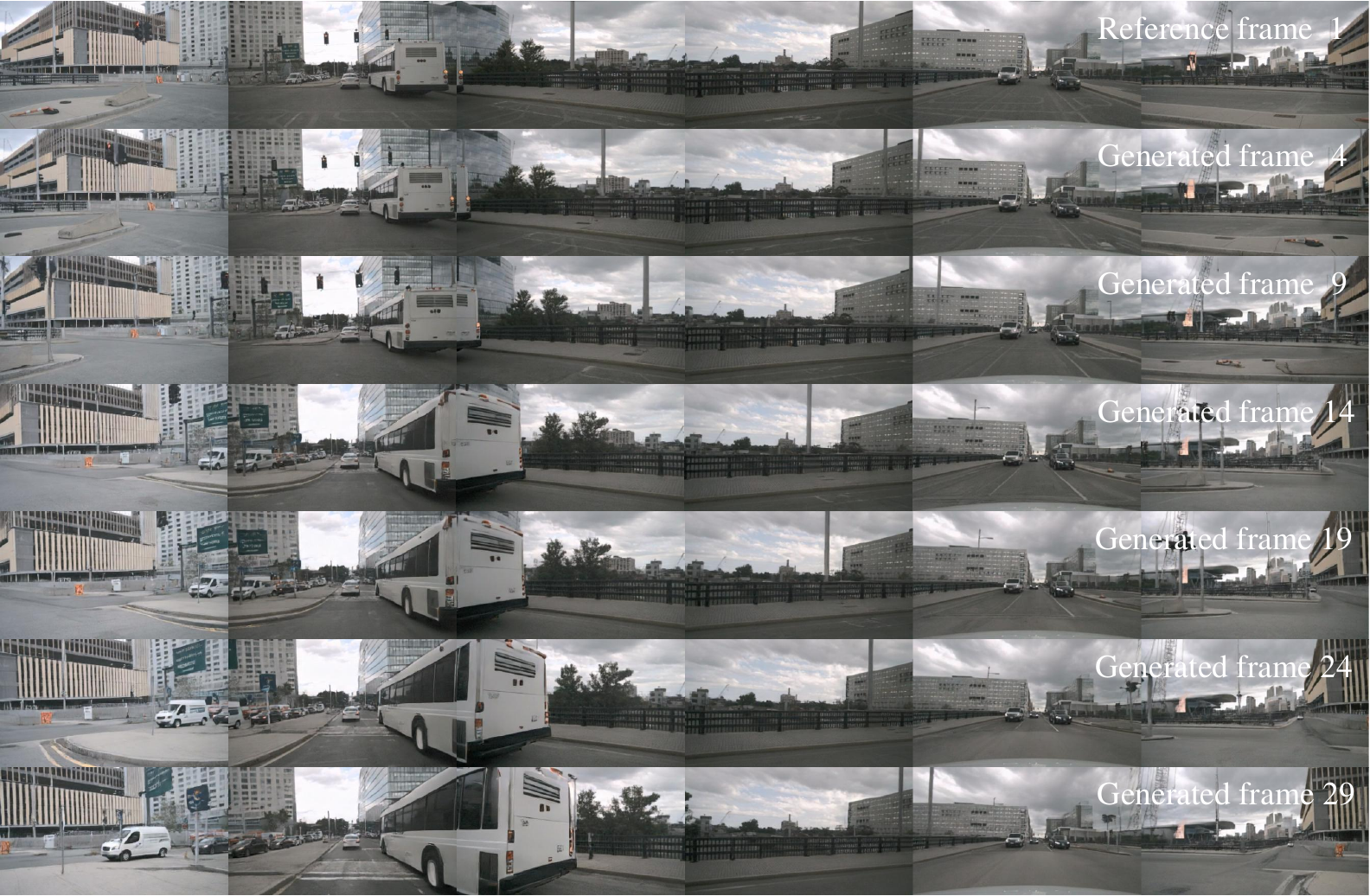}
    \caption{\textbf{Qualitative Results of Video Prediction. }We produce this example using three reference frames. The first line is the first reference frame and the following lines are the predicted frames. Our method demonstrates strong temporal consistency in the video prediction task.
    }
    \label{fig:ablat_ref3gen}
\end{figure}

\subsection{Experiment Results}
\subsubsection{Comparison}
\textbf{Generation Tasks. }Following the common evaluation protocols, we report quantitative metrics on the nuScenes validation set, shown in Table~\ref{tab:main}. Our model demonstrates exceptional performance compared to previous SOTA methods DiVE \cite{jiang2024dive} and UniMLVG \cite{unimlvg}. Specifically, our approach achieves significant improvements of 34.48\% in Fréchet Inception Distance (FID) and 61.21\% in Fréchet Video Distance (FVD) relative to the second-best method. Additionally, our model can generate high-quality videos with durations up to 20 seconds. Regarding condition consistency, our approach outperforms competing methods on mAP of object detection ($\text{mAP}_{obj}$) and IoU of road ($\text{IoU}_r$) of. It ranks second in IoU of vehicle ($\text{IoU}_v$), performing marginally below UniMLVG in this particular metric.


\textbf{STORM-VAE Results. }We provide the visualization of the depth maps of the generative images in Figure \ref{fig:vis_depth}. We put the more detailed evaluation and discussion in the Appendix.

\subsection{Ablation Study}

\begin{figure}[h]
    \centering
    \captionsetup[subfloat]{skip=0.01cm} 
    \subfloat{\includegraphics[width = 0.99\linewidth]{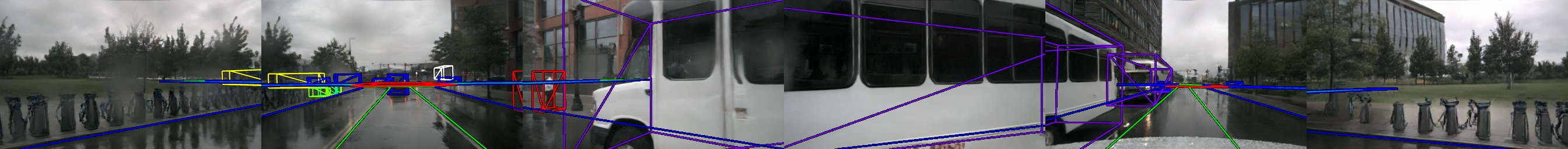}} 
    \vspace{-0.05cm}
    \subfloat{\includegraphics[width = 0.99\linewidth]{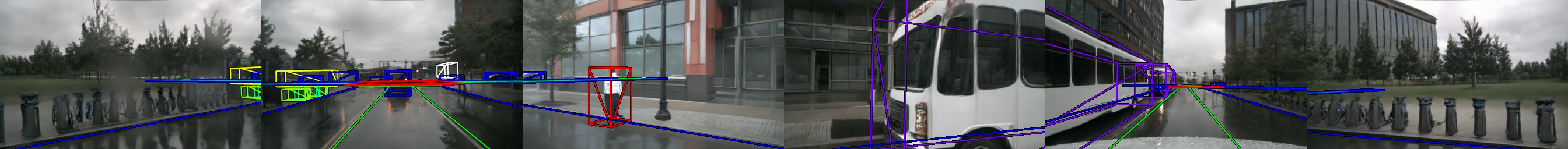}}
    \vspace{0.2cm}
    
    \subfloat{\includegraphics[width = 0.99\linewidth]{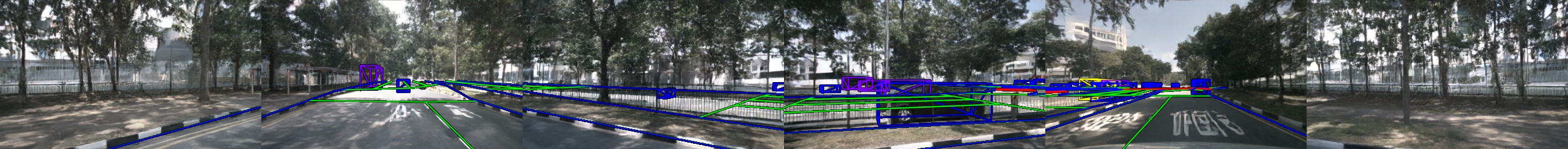}}
    \vspace{-0.05cm}
    \subfloat{\includegraphics[width = 0.99\linewidth]{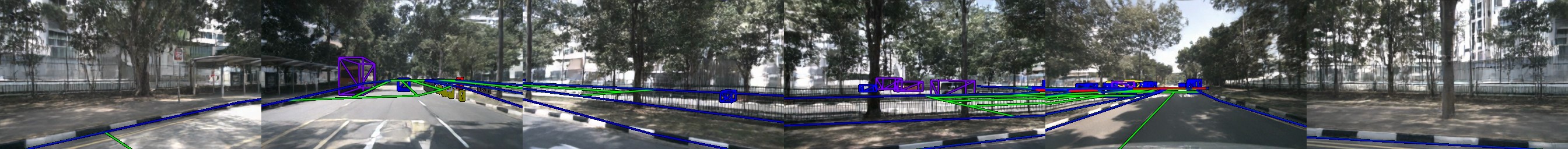}} 
    \vspace{0.2cm}
    
    \subfloat{\includegraphics[width = 0.99\linewidth]{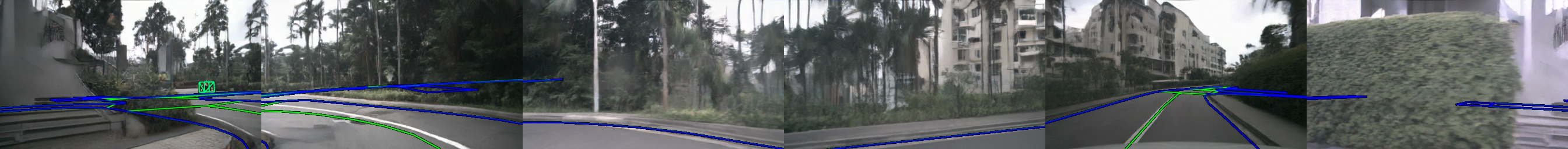}}
    \vspace{-0.05cm}
    \subfloat{\includegraphics[width = 0.99\linewidth]{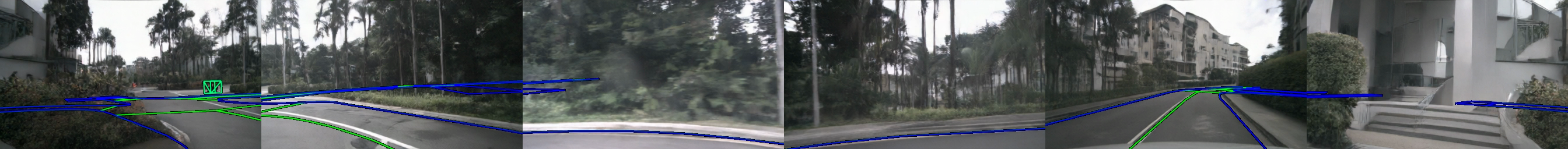}} 
    \vspace{0.2cm}

    \subfloat{\includegraphics[width = 0.99\linewidth]{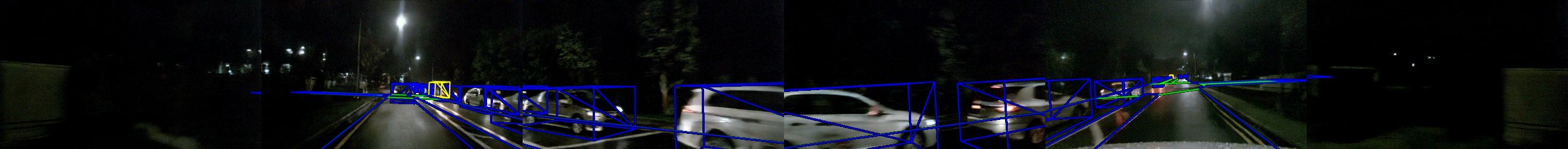}}
    \vspace{-0.05cm}
    \subfloat{\includegraphics[width = 0.99\linewidth]{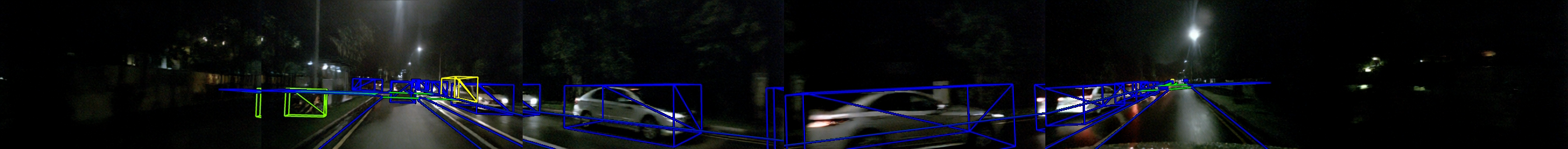}} 
    
    \vspace{0.2cm}

    \subfloat{\includegraphics[width = 0.99\linewidth]{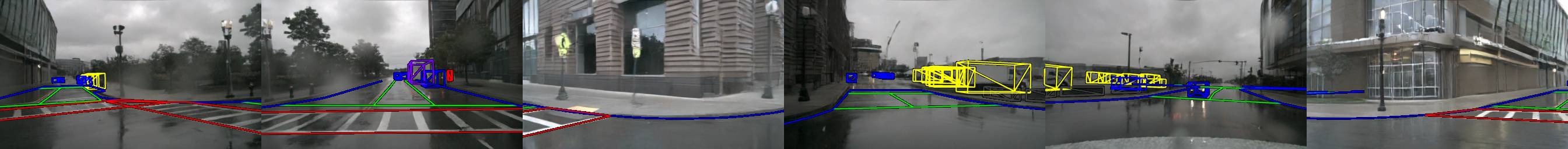}}
    \vspace{-0.05cm}
    \subfloat{\includegraphics[width = 0.99\linewidth]{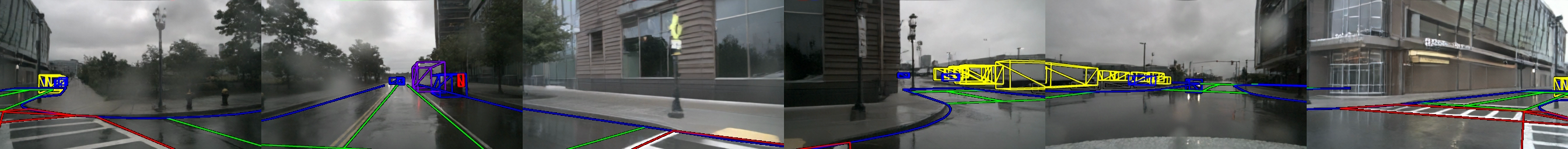}}

    \caption{\textbf{Qualitative Results of Video Generation. }We provide the examples generated with the conditons only, without any reference frame. For each scene, we list the 1st frame in the first line and the 10th frame in the second line. The bounding boxes and road maps are overlapping over the generative images. The object in the bounding boxes with the same color are should be of the same class. For example, cars should be generated in the blue 3D bounding boxes.
    }
    \label{fig:ablat_norefgen}
\end{figure}

\textbf{Number of Reference Frames. }
The number of reference frames represents different types of tasks in the generative model. Without reference frames, the model conducts pure video generation, producing content based solely on conditional inputs. On the contrary, the model perform video prediction when the reference frames are given. We present qualitative results in Figures \ref{fig:ablat_ref3gen} and~\ref{fig:ablat_norefgen}, with quantitative evaluations in Table \ref{tab:ref_frame}.
As shown in the table, the FVD score is steadily improved as the number of reference frames increases, indicating that additional reference frames provide richer temporal information from the ground truth, thereby improving temporal consistency. Conversely, when reference frames are provided, the model performs video prediction. For more results, please refer to the Appendix.

\textbf{Effect of STORM-VAE. }Table~\ref{tab:storm_vae} demonstrates that our STORM-VAE significantly improves generation quality over the standard VAE baseline. Specifically, STORM-VAE yields a 15.38\% reduction in Fréchet Inception Distance (FID) and a 34.97\% decrease in Fréchet Video Distance (FVD), indicating substantial enhancements in both image and video generation quality. Furthermore, Figure~\ref{fig:teaser} illustrates that our model accelerates convergence compared to the baseline. To ensure fair evaluation in this ablation study, we compare models trained for the same number of steps.

\section{Conclusion}
We introduce \ours, a novel framework that unifies long-sequence, multi-view video generation with dynamic 4D scene reconstruction. Our approach extends the traditional VAE architecture by incorporating a Gaussian Splatting Decoder, namely STORM-VAE. This design not only enables high-quality 4D scene reconstruction but also substantially enhances representation learning, thereby improving the generative capabilities of our downstream diffusion model. Leveraging the pre-trained STORM-VAE, we train \ours~using multiple datasets and support various conditioning types across diverse generative tasks. Experimental results demonstrate that \ours~surpasses SOTA methods, particularly in image quality and temporal coherence. Furthermore, the Gaussian Splatting Decoder directly estimates absolute depth through neural rendering, providing richer 3D structural information than previous approaches.

\bibliography{iclr2025_conference}

\begin{thebibliography}{54}
\providecommand{\natexlab}[1]{#1}
\providecommand{\url}[1]{\texttt{#1}}
\expandafter\ifx\csname urlstyle\endcsname\relax
  \providecommand{\doi}[1]{doi: #1}\else
  \providecommand{\doi}{doi: \begingroup \urlstyle{rm}\Url}\fi

\bibitem[Caesar et~al.(2020)Caesar, Bankiti, Lang, Vora, Liong, Xu, Krishnan, Pan, Baldan, and Beijbom]{caesar2020nuscenes}
Holger Caesar, Varun Bankiti, Alex~H Lang, Sourabh Vora, Venice~Erin Liong, Qiang Xu, Anush Krishnan, Yu~Pan, Giancarlo Baldan, and Oscar Beijbom.
\newblock nuscenes: A multimodal dataset for autonomous driving.
\newblock In \emph{Proceedings of the IEEE/CVF conference on computer vision and pattern recognition}, pp.\  11621--11631, 2020.

\bibitem[Chen et~al.(2024)Chen, Wu, Liu, Guo, Ni, Xia, and Xia]{unimlvg}
Rui Chen, Zehuan Wu, Yichen Liu, Yuxin Guo, Jingcheng Ni, Haifeng Xia, and Siyu Xia.
\newblock Unimlvg: Unified framework for multi-view long video generation with comprehensive control capabilities for autonomous driving.
\newblock \emph{arXiv preprint arXiv:2412.04842}, 2024.

\bibitem[Deja et~al.(2023)Deja, Trzci{\'n}ski, and Tomczak]{deja2023learning}
Kamil Deja, Tomasz Trzci{\'n}ski, and Jakub~M Tomczak.
\newblock Learning data representations with joint diffusion models.
\newblock In \emph{Joint European Conference on Machine Learning and Knowledge Discovery in Databases}, pp.\  543--559. Springer, 2023.

\bibitem[Esser et~al.(2024)Esser, Kulal, Blattmann, Entezari, M{\"u}ller, Saini, Levi, Lorenz, Sauer, Boesel, et~al.]{sd35}
Patrick Esser, Sumith Kulal, Andreas Blattmann, Rahim Entezari, Jonas M{\"u}ller, Harry Saini, Yam Levi, Dominik Lorenz, Axel Sauer, Frederic Boesel, et~al.
\newblock Scaling rectified flow transformers for high-resolution image synthesis.
\newblock In \emph{Forty-first international conference on machine learning}, 2024.

\bibitem[Fuest et~al.(2024)Fuest, Ma, Gui, Schusterbauer, Hu, and Ommer]{fuest2024diffusion}
Michael Fuest, Pingchuan Ma, Ming Gui, Johannes Schusterbauer, Vincent~Tao Hu, and Bjorn Ommer.
\newblock Diffusion models and representation learning: A survey.
\newblock \emph{arXiv preprint arXiv:2407.00783}, 2024.

\bibitem[Gao et~al.(2023)Gao, Chen, Xie, Hong, Li, Yeung, and Xu]{magicdrive}
Ruiyuan Gao, Kai Chen, Enze Xie, Lanqing Hong, Zhenguo Li, Dit-Yan Yeung, and Qiang Xu.
\newblock Magicdrive: Street view generation with diverse 3d geometry control.
\newblock \emph{arXiv preprint arXiv:2310.02601}, 2023.

\bibitem[Gao et~al.(2024{\natexlab{a}})Gao, Chen, Li, Hong, Li, and Xu]{gao2024magicdrive3d}
Ruiyuan Gao, Kai Chen, Zhihao Li, Lanqing Hong, Zhenguo Li, and Qiang Xu.
\newblock Magicdrive3d: Controllable 3d generation for any-view rendering in street scenes.
\newblock \emph{arXiv preprint arXiv:2405.14475}, 2024{\natexlab{a}}.

\bibitem[Gao et~al.(2024{\natexlab{b}})Gao, Chen, Xiao, Hong, Li, and Xu]{magicdrivev2}
Ruiyuan Gao, Kai Chen, Bo~Xiao, Lanqing Hong, Zhenguo Li, and Qiang Xu.
\newblock Magicdrive-v2: High-resolution long video generation for autonomous driving with adaptive control.
\newblock \emph{arXiv preprint arXiv:2411.13807}, 2024{\natexlab{b}}.

\bibitem[Gao et~al.(2024{\natexlab{c}})Gao, Yang, Chen, Chitta, Qiu, Geiger, Zhang, and Li]{vista}
Shenyuan Gao, Jiazhi Yang, Li~Chen, Kashyap Chitta, Yihang Qiu, Andreas Geiger, Jun Zhang, and Hongyang Li.
\newblock Vista: A generalizable driving world model with high fidelity and versatile controllability.
\newblock \emph{Advances in Neural Information Processing Systems}, 37:\penalty0 91560--91596, 2024{\natexlab{c}}.

\bibitem[Gao et~al.(2025)Gao, Guo, Hoang, Huang, Jiang, Kong, Li, Li, Li, Li, et~al.]{seedance}
Yu~Gao, Haoyuan Guo, Tuyen Hoang, Weilin Huang, Lu~Jiang, Fangyuan Kong, Huixia Li, Jiashi Li, Liang Li, Xiaojie Li, et~al.
\newblock Seedance 1.0: Exploring the boundaries of video generation models.
\newblock \emph{arXiv preprint arXiv:2506.09113}, 2025.

\bibitem[Hassan et~al.(2025)Hassan, Stapf, Rahimi, Rezende, Haghighi, Br{\"u}ggemann, Katircioglu, Zhang, Chen, Saha, et~al.]{hassan2025gem}
Mariam Hassan, Sebastian Stapf, Ahmad Rahimi, Pedro Rezende, Yasaman Haghighi, David Br{\"u}ggemann, Isinsu Katircioglu, Lin Zhang, Xiaoran Chen, Suman Saha, et~al.
\newblock Gem: A generalizable ego-vision multimodal world model for fine-grained ego-motion, object dynamics, and scene composition control.
\newblock In \emph{Proceedings of the Computer Vision and Pattern Recognition Conference}, pp.\  22404--22415, 2025.

\bibitem[Heusel et~al.(2017)Heusel, Ramsauer, Unterthiner, Nessler, and Hochreiter]{heusel2017gans}
Martin Heusel, Hubert Ramsauer, Thomas Unterthiner, Bernhard Nessler, and Sepp Hochreiter.
\newblock Gans trained by a two time-scale update rule converge to a local nash equilibrium.
\newblock \emph{Advances in neural information processing systems}, 30, 2017.

\bibitem[Hong et~al.(2022)Hong, Ding, Zheng, Liu, and Tang]{cogvideo}
Wenyi Hong, Ming Ding, Wendi Zheng, Xinghan Liu, and Jie Tang.
\newblock Cogvideo: Large-scale pretraining for text-to-video generation via transformers.
\newblock \emph{arXiv preprint arXiv:2205.15868}, 2022.

\bibitem[Jiang et~al.(2024)Jiang, Hong, Zhou, Ma, Hu, Zhou, Xiang, Liu, Yu, Sun, et~al.]{jiang2024dive}
Junpeng Jiang, Gangyi Hong, Lijun Zhou, Enhui Ma, Hengtong Hu, Xia Zhou, Jie Xiang, Fan Liu, Kaicheng Yu, Haiyang Sun, et~al.
\newblock Dive: Dit-based video generation with enhanced control.
\newblock \emph{arXiv preprint arXiv:2409.01595}, 2024.

\bibitem[Kerbl et~al.(2023)Kerbl, Kopanas, Leimk{\"u}hler, and Drettakis]{3dgs}
Bernhard Kerbl, Georgios Kopanas, Thomas Leimk{\"u}hler, and George Drettakis.
\newblock 3d gaussian splatting for real-time radiance field rendering.
\newblock \emph{ACM Trans. Graph.}, 42\penalty0 (4):\penalty0 139--1, 2023.

\bibitem[Kim et~al.(2021)Kim, Philion, Torralba, and Fidler]{drivegan}
Seung~Wook Kim, Jonah Philion, Antonio Torralba, and Sanja Fidler.
\newblock Drivegan: Towards a controllable high-quality neural simulation.
\newblock In \emph{Proceedings of the IEEE/CVF Conference on Computer Vision and Pattern Recognition}, pp.\  5820--5829, 2021.

\bibitem[Kong et~al.(2024)Kong, Tian, Zhang, Min, Dai, Zhou, Xiong, Li, Wu, Zhang, and Others]{hunyuanvideo}
Weijie Kong, Qi~Tian, Zijian Zhang, Rox Min, Zuozhuo Dai, Jin Zhou, Jiangfeng Xiong, Xin Li, Bo~Wu, Jianwei Zhang, and Others.
\newblock Hunyuanvideo: A systematic framework for large video generative models.
\newblock \emph{arXiv preprint arXiv:2412.03603}, 2024.

\bibitem[Labs et~al.(2025)Labs, Batifol, Blattmann, Boesel, Consul, Diagne, Dockhorn, English, English, Esser, Kulal, Lacey, Levi, Li, Lorenz, Müller, Podell, Rombach, Saini, Sauer, and Smith]{flux1}
Black~Forest Labs, Stephen Batifol, Andreas Blattmann, Frederic Boesel, Saksham Consul, Cyril Diagne, Tim Dockhorn, Jack English, Zion English, Patrick Esser, Sumith Kulal, Kyle Lacey, Yam Levi, Cheng Li, Dominik Lorenz, Jonas Müller, Dustin Podell, Robin Rombach, Harry Saini, Axel Sauer, and Luke Smith.
\newblock Flux.1 kontext: Flow matching for in-context image generation and editing in latent space, 2025.
\newblock URL \url{https://arxiv.org/abs/2506.15742}.

\bibitem[Leng et~al.(2025)Leng, Singh, Hou, Xing, Xie, and Zheng]{repae}
Xingjian Leng, Jaskirat Singh, Yunzhong Hou, Zhenchang Xing, Saining Xie, and Liang Zheng.
\newblock Repa-e: Unlocking vae for end-to-end tuning with latent diffusion transformers.
\newblock \emph{arXiv preprint arXiv:2504.10483}, 2025.

\bibitem[Li et~al.(2025)Li, Guo, Liu, Zou, Ding, Chen, Zhu, Tan, Zhang, Wang, et~al.]{li2025uniscene}
Bohan Li, Jiazhe Guo, Hongsi Liu, Yingshuang Zou, Yikang Ding, Xiwu Chen, Hu~Zhu, Feiyang Tan, Chi Zhang, Tiancai Wang, et~al.
\newblock Uniscene: Unified occupancy-centric driving scene generation.
\newblock In \emph{Proceedings of the Computer Vision and Pattern Recognition Conference}, pp.\  11971--11981, 2025.

\bibitem[Li et~al.(2024)Li, Zhang, Lin, Xiong, Long, Deng, Zhang, Liu, Huang, Xiao, et~al.]{hunyuan-dit}
Zhimin Li, Jianwei Zhang, Qin Lin, Jiangfeng Xiong, Yanxin Long, Xinchi Deng, Yingfang Zhang, Xingchao Liu, Minbin Huang, Zedong Xiao, et~al.
\newblock Hunyuan-dit: A powerful multi-resolution diffusion transformer with fine-grained chinese understanding.
\newblock \emph{arXiv preprint arXiv:2405.08748}, 2024.

\bibitem[Li et~al.(2022)Li, Wang, Li, Xie, Sima, Lu, Qiao, and Dai]{li2022bevformer}
Zhiqi Li, Wenhai Wang, Hongyang Li, Enze Xie, Chonghao Sima, Tong Lu, Yu~Qiao, and Jifeng Dai.
\newblock Bevformer: Learning bird’s-eye-view representation from multi-camera images via spatiotemporal transformers.
\newblock \emph{arXiv preprint arXiv:2203.17270}, 2022.

\bibitem[Liang et~al.(2025)Liang, Zhang, Zhou, Tu, Feng, Li, Zhang, Du, Tan, and Bai]{unifuture}
Dingkang Liang, Dingyuan Zhang, Xin Zhou, Sifan Tu, Tianrui Feng, Xiaofan Li, Yumeng Zhang, Mingyang Du, Xiao Tan, and Xiang Bai.
\newblock Seeing the future, perceiving the future: A unified driving world model for future generation and perception.
\newblock \emph{arXiv preprint arXiv:2503.13587}, 2025.

\bibitem[Liu et~al.(2022)Liu, Gong, and Liu]{liu2022flow}
Xingchao Liu, Chengyue Gong, and Qiang Liu.
\newblock Flow straight and fast: Learning to generate and transfer data with rectified flow.
\newblock \emph{arXiv preprint arXiv:2209.03003}, 2022.

\bibitem[Mei et~al.(2024)Mei, Hu, Yang, Wen, Yang, Wei, Ma, Dou, Shi, and Liu]{dreamforge}
Jianbiao Mei, Tao Hu, Xuemeng Yang, Licheng Wen, Yu~Yang, Tiantian Wei, Yukai Ma, Min Dou, Botian Shi, and Yong Liu.
\newblock Dreamforge: Motion-aware autoregressive video generation for multi-view driving scenes.
\newblock \emph{arXiv preprint arXiv:2409.04003}, 2024.

\bibitem[Ni et~al.(2025{\natexlab{a}})Ni, Zhao, Wang, Zhu, Qin, Huang, Liu, Chen, Wang, Zhang, et~al.]{ni2025recondreamer}
Chaojun Ni, Guosheng Zhao, Xiaofeng Wang, Zheng Zhu, Wenkang Qin, Guan Huang, Chen Liu, Yuyin Chen, Yida Wang, Xueyang Zhang, et~al.
\newblock Recondreamer: Crafting world models for driving scene reconstruction via online restoration.
\newblock In \emph{Proceedings of the Computer Vision and Pattern Recognition Conference}, pp.\  1559--1569, 2025{\natexlab{a}}.

\bibitem[Ni et~al.(2025{\natexlab{b}})Ni, Guo, Liu, Chen, Lu, and Wu]{maskgwm}
Jingcheng Ni, Yuxin Guo, Yichen Liu, Rui Chen, Lewei Lu, and Zehuan Wu.
\newblock Maskgwm: A generalizable driving world model with video mask reconstruction.
\newblock In \emph{Proceedings of the Computer Vision and Pattern Recognition Conference}, pp.\  22381--22391, 2025{\natexlab{b}}.

\bibitem[Peebles \& Xie(2023)Peebles and Xie]{dit}
William Peebles and Saining Xie.
\newblock Scalable diffusion models with transformers.
\newblock In \emph{Proceedings of the IEEE/CVF international conference on computer vision}, pp.\  4195--4205, 2023.

\bibitem[Peng et~al.(2025)Peng, Zheng, Shen, Young, Guo, Wang, Xu, Liu, Jiang, Li, Wang, Ye, Ren, Ma, Liang, Lian, Wu, Zhong, Li, Gong, Lei, Cheng, Zhang, Li, Zhang, Hu, Huang, Wang, Zhao, Wang, Wei, and You]{opensora2}
Xiangyu Peng, Zangwei Zheng, Chenhui Shen, Tom Young, Xinying Guo, Binluo Wang, Hang Xu, Hongxin Liu, Mingyan Jiang, Wenjun Li, Yuhui Wang, Anbang Ye, Gang Ren, Qianran Ma, Wanying Liang, Xiang Lian, Xiwen Wu, Yuting Zhong, Zhuangyan Li, Chaoyu Gong, Guojun Lei, Leijun Cheng, Limin Zhang, Minghao Li, Ruijie Zhang, Silan Hu, Shijie Huang, Xiaokang Wang, Yuanheng Zhao, Yuqi Wang, Ziang Wei, and Yang You.
\newblock Open-sora 2.0: Training a commercial-level video generation model in 200k.
\newblock \emph{arXiv preprint arXiv:2503.09642}, 2025.

\bibitem[Pernias et~al.(2023)Pernias, Rampas, Richter, Pal, and Aubreville]{pernias2023wurstchen}
Pablo Pernias, Dominic Rampas, Mats~L Richter, Christopher~J Pal, and Marc Aubreville.
\newblock W{\"u}rstchen: An efficient architecture for large-scale text-to-image diffusion models.
\newblock \emph{arXiv preprint arXiv:2306.00637}, 2023.

\bibitem[Ren et~al.(2025)Ren, Lu, Cao, Gao, Huang, Sabour, Shen, Pfaff, Wu, Chen, et~al.]{cosmos-drive-dreams}
Xuanchi Ren, Yifan Lu, Tianshi Cao, Ruiyuan Gao, Shengyu Huang, Amirmojtaba Sabour, Tianchang Shen, Tobias Pfaff, Jay~Zhangjie Wu, Runjian Chen, et~al.
\newblock Cosmos-drive-dreams: Scalable synthetic driving data generation with world foundation models.
\newblock \emph{arXiv preprint arXiv:2506.09042}, 2025.

\bibitem[Rombach et~al.(2022)Rombach, Blattmann, Lorenz, Esser, and Ommer]{stable-diffusion}
Robin Rombach, Andreas Blattmann, Dominik Lorenz, Patrick Esser, and Bj{\"o}rn Ommer.
\newblock High-resolution image synthesis with latent diffusion models.
\newblock In \emph{Proceedings of the IEEE/CVF conference on computer vision and pattern recognition}, pp.\  10684--10695, 2022.

\bibitem[Russell et~al.(2025)Russell, Hu, Bertoni, Fedoseev, Shotton, Arani, and Corrado]{gaia2}
Lloyd Russell, Anthony Hu, Lorenzo Bertoni, George Fedoseev, Jamie Shotton, Elahe Arani, and Gianluca Corrado.
\newblock Gaia-2: A controllable multi-view generative world model for autonomous driving.
\newblock \emph{arXiv preprint arXiv:2503.20523}, 2025.

\bibitem[Sun et~al.(2020)Sun, Kretzschmar, Dotiwalla, Chouard, Patnaik, Tsui, Guo, Zhou, Chai, Caine, et~al.]{sun2020scalability}
Pei Sun, Henrik Kretzschmar, Xerxes Dotiwalla, Aurelien Chouard, Vijaysai Patnaik, Paul Tsui, James Guo, Yin Zhou, Yuning Chai, Benjamin Caine, et~al.
\newblock Scalability in perception for autonomous driving: Waymo open dataset.
\newblock In \emph{Proceedings of the IEEE/CVF conference on computer vision and pattern recognition}, pp.\  2446--2454, 2020.

\bibitem[Tian et~al.(2023)Tian, Tao, Dai, Li, Li, Lu, Wang, Li, Huang, and Zhu]{addp}
Changyao Tian, Chenxin Tao, Jifeng Dai, Hao Li, Ziheng Li, Lewei Lu, Xiaogang Wang, Hongsheng Li, Gao Huang, and Xizhou Zhu.
\newblock Addp: Learning general representations for image recognition and generation with alternating denoising diffusion process.
\newblock \emph{arXiv preprint arXiv:2306.05423}, 2023.

\bibitem[Unterthiner et~al.(2018)Unterthiner, Van~Steenkiste, Kurach, Marinier, Michalski, and Gelly]{unterthiner2018towards}
Thomas Unterthiner, Sjoerd Van~Steenkiste, Karol Kurach, Raphael Marinier, Marcin Michalski, and Sylvain Gelly.
\newblock Towards accurate generative models of video: A new metric \& challenges.
\newblock \emph{arXiv preprint arXiv:1812.01717}, 2018.

\bibitem[Wei et~al.(2025)Wei, Li, and Liu]{wei2025omni}
Dongxu Wei, Zhiqi Li, and Peidong Liu.
\newblock Omni-scene: Omni-gaussian representation for ego-centric sparse-view scene reconstruction.
\newblock In \emph{Proceedings of the Computer Vision and Pattern Recognition Conference}, pp.\  22317--22327, 2025.

\bibitem[Wilson et~al.(2023)Wilson, Qi, Agarwal, Lambert, Singh, Khandelwal, Pan, Kumar, Hartnett, Pontes, et~al.]{wilson2023argoverse}
Benjamin Wilson, William Qi, Tanmay Agarwal, John Lambert, Jagjeet Singh, Siddhesh Khandelwal, Bowen Pan, Ratnesh Kumar, Andrew Hartnett, Jhony~Kaesemodel Pontes, et~al.
\newblock Argoverse 2: Next generation datasets for self-driving perception and forecasting.
\newblock \emph{arXiv preprint arXiv:2301.00493}, 2023.

\bibitem[Wu et~al.(2024{\natexlab{a}})Wu, Guo, Tang, Huang, Wang, Chen, and Ding]{drivescape}
Wei Wu, Xi~Guo, Weixuan Tang, Tingxuan Huang, Chiyu Wang, Dongyue Chen, and Chenjing Ding.
\newblock Drivescape: Towards high-resolution controllable multi-view driving video generation.
\newblock \emph{arXiv preprint arXiv:2409.05463}, 2024{\natexlab{a}}.

\bibitem[Wu et~al.(2024{\natexlab{b}})Wu, Ni, Wang, Guo, Chen, Lu, Dai, and Xiong]{holodrive}
Zehuan Wu, Jingcheng Ni, Xiaodong Wang, Yuxin Guo, Rui Chen, Lewei Lu, Jifeng Dai, and Yuwen Xiong.
\newblock Holodrive: Holistic 2d-3d multi-modal street scene generation for autonomous driving.
\newblock \emph{arXiv preprint arXiv:2412.01407}, 2024{\natexlab{b}}.

\bibitem[Xie et~al.(2025)Xie, Liu, Wang, Cao, and Zhang]{xie2025glad}
Bin Xie, Yingfei Liu, Tiancai Wang, Jiale Cao, and Xiangyu Zhang.
\newblock Glad: A streaming scene generator for autonomous driving.
\newblock \emph{arXiv preprint arXiv:2503.00045}, 2025.

\bibitem[Yan et~al.(2025)Yan, Wu, Han, Jiang, Zhou, Zhan, Xu, and Shen]{yan2025drivingsphere}
Tianyi Yan, Dongming Wu, Wencheng Han, Junpeng Jiang, Xia Zhou, Kun Zhan, Cheng-zhong Xu, and Jianbing Shen.
\newblock Drivingsphere: Building a high-fidelity 4d world for closed-loop simulation.
\newblock In \emph{Proceedings of the Computer Vision and Pattern Recognition Conference}, pp.\  27531--27541, 2025.

\bibitem[Yang et~al.(2025)Yang, Huang, Chen, Wang, Li, You, Igl, Sharma, Karkus, Xu, Ivanovic, Wang, and Pavone]{yang2025storm}
Jiawei Yang, Jiahui Huang, Yuxiao Chen, Yan Wang, Boyi Li, Yurong You, Maximilian Igl, Apoorva Sharma, Peter Karkus, Danfei Xu, Boris Ivanovic, Yue Wang, and Marco Pavone.
\newblock Storm: Spatio-temporal reconstruction model for large-scale outdoor scenes.
\newblock \emph{arXiv preprint arXiv:2501.00602}, 2025.

\bibitem[Yang et~al.(2024{\natexlab{a}})Yang, Gao, Qiu, Chen, Li, Dai, Chitta, Wu, Zeng, Luo, Zhang, Geiger, Qiao, and Li]{genad}
Jiazhi Yang, Shenyuan Gao, Yihang Qiu, Li~Chen, Tianyu Li, Bo~Dai, Kashyap Chitta, Penghao Wu, Jia Zeng, Ping Luo, Jun Zhang, Andreas Geiger, Yu~Qiao, and Hongyang Li.
\newblock Generalized predictive model for autonomous driving.
\newblock In \emph{Proceedings of the IEEE/CVF Conference on Computer Vision and Pattern Recognition (CVPR)}, pp.\  14662--14672, June 2024{\natexlab{a}}.

\bibitem[Yang et~al.(2024{\natexlab{b}})Yang, Gao, Qiu, Chen, Li, Dai, Chitta, Wu, Zeng, Luo, et~al.]{yang2024generalized}
Jiazhi Yang, Shenyuan Gao, Yihang Qiu, Li~Chen, Tianyu Li, Bo~Dai, Kashyap Chitta, Penghao Wu, Jia Zeng, Ping Luo, et~al.
\newblock Generalized predictive model for autonomous driving.
\newblock In \emph{Proceedings of the IEEE/CVF Conference on Computer Vision and Pattern Recognition}, pp.\  14662--14672, 2024{\natexlab{b}}.

\bibitem[Yang et~al.(2024{\natexlab{c}})Yang, Kang, Huang, Zhao, Xu, Feng, and Zhao]{yang2024depth}
Lihe Yang, Bingyi Kang, Zilong Huang, Zhen Zhao, Xiaogang Xu, Jiashi Feng, and Hengshuang Zhao.
\newblock Depth anything v2.
\newblock \emph{Advances in Neural Information Processing Systems}, 37:\penalty0 21875--21911, 2024{\natexlab{c}}.

\bibitem[Yang et~al.(2022)Yang, Shih, Fu, Zhao, and Ji]{yang2022your}
Xiulong Yang, Sheng-Min Shih, Yinlin Fu, Xiaoting Zhao, and Shihao Ji.
\newblock Your vit is secretly a hybrid discriminative-generative diffusion model.
\newblock \emph{arXiv preprint arXiv:2208.07791}, 2022.

\bibitem[Yang et~al.(2024{\natexlab{d}})Yang, Teng, Zheng, Ding, Huang, Xu, Yang, Hong, Zhang, Feng, et~al.]{cogvideox}
Zhuoyi Yang, Jiayan Teng, Wendi Zheng, Ming Ding, Shiyu Huang, Jiazheng Xu, Yuanming Yang, Wenyi Hong, Xiaohan Zhang, Guanyu Feng, et~al.
\newblock Cogvideox: Text-to-video diffusion models with an expert transformer.
\newblock \emph{arXiv preprint arXiv:2408.06072}, 2024{\natexlab{d}}.

\bibitem[Yao et~al.(2025)Yao, Yang, and Wang]{vavae}
Jingfeng Yao, Bin Yang, and Xinggang Wang.
\newblock Reconstruction vs. generation: Taming optimization dilemma in latent diffusion models.
\newblock In \emph{Proceedings of the Computer Vision and Pattern Recognition Conference}, pp.\  15703--15712, 2025.

\bibitem[Yu et~al.(2025)Yu, Kwak, Jang, Jeong, Huang, Shin, and Xie]{repa}
Sihyun Yu, Sangkyung Kwak, Huiwon Jang, Jongheon Jeong, Jonathan Huang, Jinwoo Shin, and Saining Xie.
\newblock Representation alignment for generation: Training diffusion transformers is easier than you think.
\newblock In \emph{International Conference on Learning Representations}, 2025.

\bibitem[Zhang et~al.(2023)Zhang, Rao, and Agrawala]{controlnet}
Lvmin Zhang, Anyi Rao, and Maneesh Agrawala.
\newblock Adding conditional control to text-to-image diffusion models.
\newblock In \emph{Proceedings of the IEEE/CVF international conference on computer vision}, pp.\  3836--3847, 2023.

\bibitem[Zhao et~al.(2025)Zhao, Wang, Zhu, Chen, Huang, Bao, and Wang]{drivedreamer2}
Guosheng Zhao, Xiaofeng Wang, Zheng Zhu, Xinze Chen, Guan Huang, Xiaoyi Bao, and Xingang Wang.
\newblock Drivedreamer-2: Llm-enhanced world models for diverse driving video generation.
\newblock In \emph{Proceedings of the AAAI Conference on Artificial Intelligence}, volume~39, pp.\  10412--10420, 2025.

\bibitem[Zheng et~al.(2024)Zheng, Peng, Yang, Shen, Li, Liu, Zhou, Li, and You]{opensora}
Zangwei Zheng, Xiangyu Peng, Tianji Yang, Chenhui Shen, Shenggui Li, Hongxin Liu, Yukun Zhou, Tianyi Li, and Yang You.
\newblock Open-sora: Democratizing efficient video production for all.
\newblock \emph{arXiv preprint arXiv:2412.20404}, 2024.

\bibitem[Zhou \& Kr{\"a}henb{\"u}hl(2022)Zhou and Kr{\"a}henb{\"u}hl]{zhou2022cross}
Brady Zhou and Philipp Kr{\"a}henb{\"u}hl.
\newblock Cross-view transformers for real-time map-view semantic segmentation.
\newblock In \emph{CVPR}, 2022.

\end{thebibliography}
\bibliographystyle{iclr2025_conference}

\newpage
\section{Appendix}


\subsection{Additional Experiments}

\begin{table}[!htb]
    \caption{\textbf{Comparison of STORM and STORM-VAE}}
\label{tab:comp:storm}
    
    \vspace{-0.2cm}
    \begin{subtable}[t]{.5\linewidth}
      \centering
        \begin{tabular}{c|cc}
            \toprule
            \textbf{Method} & {PSNR} $\uparrow$ & {D-RMSE} $\downarrow$   \\
            \midrule
            STORM & 20.89 & 5.52 \\
            \midrule
            STORM-VAE  & 21.18 & 4.55 \\
            \bottomrule
        \end{tabular}
        \caption{\textbf{Comparison of Reconstruction. } We extend the original STORM to a 6-view rendering model and evaluate the performance on NuScene. Our STORM-VAE also slightly outperforms the STORM in the reconstruction task.}
        \label{tab:comp:storm:recon}
    \end{subtable}%
    \hspace{0.2cm}
    \begin{subtable}[t]{.5\linewidth}
      \centering
        \begin{tabular}{c|cc}
            \toprule
            \textbf{Method} & {AbsRel} $\downarrow$ & $\bm {\delta_1}$ $\uparrow$   \\
            \midrule
            UniMLVG + STORM & 30.825 & 49.7  \\
            \midrule
            \ours & 16.05 & 49.7  \\ 
            \bottomrule
        \end{tabular}
        \caption{\textbf{Comparison of Zero-shot Depth Estimate. } We evaluate the performance of our models on depth estimation in the generation results. We use the pesudo-groundtruth produced by Depth Anything V2. 
        }
        \label{tab:comp:storm:gen}
    \end{subtable}
    
\end{table}

\subsubsection{Video Results}
We provide a video in the supplementary material for better visualization.

\subsubsection{Comparison of STORM and STORM-VAE. }We evaluate the performance of STORM-VAE in comparison to STORM, as illustrated in Table \ref{tab:comp:storm}. Specifically, Table \ref{tab:comp:storm:recon} demonstrates STORM-VAE's reconstruction capabilities relative to STORM. For quantitative assessment, we evaluate the reconstructed images and depth maps of STORM-VAE on the nuScenes dataset using two metrics: Peak Signal-to-Noise Ratio (PSNR) for image quality and Depth Root Mean Square Error (D-RMSE) for depth accuracy. Our experimental results demonstrate that STORM-VAE even slightly exceeds its performance. 

\begin{figure}[t]
    \centering 
    \captionsetup[subfloat]{position=bottom, labelformat=empty, font=tiny, justification=justified, singlelinecheck=false}
    \captionsetup[subfloat]{skip=0.01cm}
    \subfloat{\includegraphics[width = 0.99\linewidth]{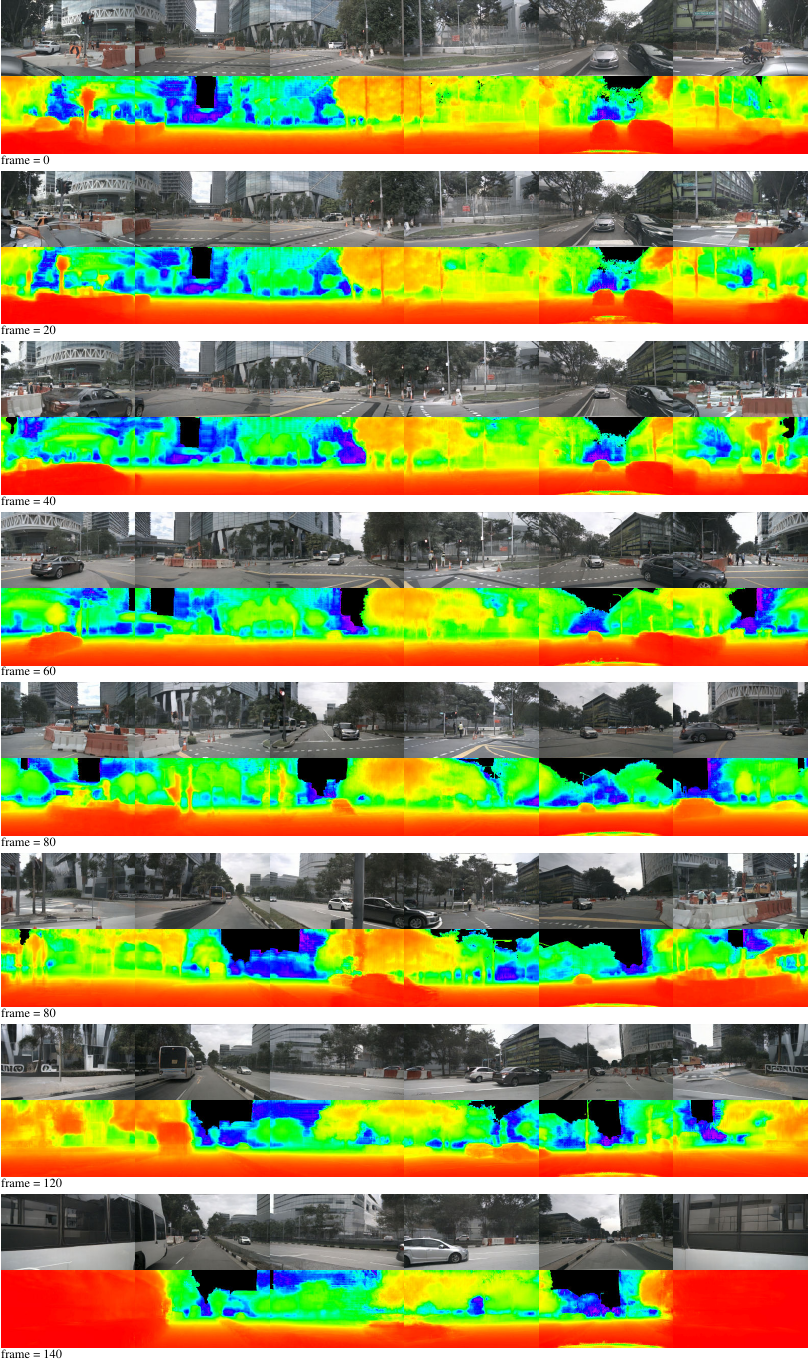}}

    \vspace{-0.4cm}
    \caption{\textbf{Qualitative results of Depth Estimation. }}
    \label{fig:vis_depth_app_0}
\end{figure}

\begin{figure}[t]
    \centering 
    \captionsetup[subfloat]{position=bottom, labelformat=empty, font=tiny, justification=justified, singlelinecheck=false}
    \captionsetup[subfloat]{skip=0.01cm}
    \subfloat{\includegraphics[width = 0.99\linewidth]{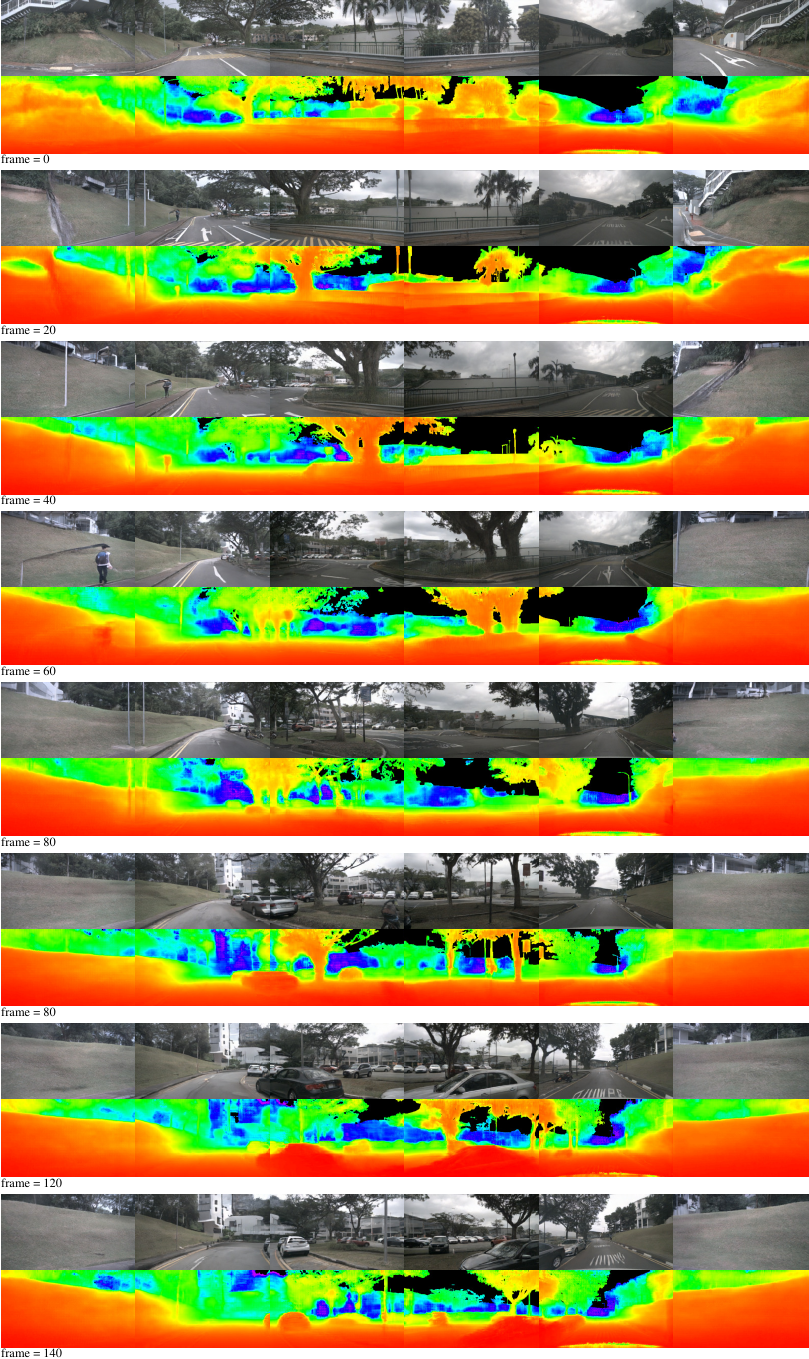}}

    \vspace{-0.4cm}
    \caption{\textbf{Qualitative results of Depth Estimation. }}
    \label{fig:vis_depth_app_1}
\end{figure}

In generation task, we compare the performance of \ours~against UniMLVG + STORM, which first employs UniMLVG to generate videos and subsequently applies STORM to reconstruct the 4D scene. During inference, we set the context timesteps equal to the target timesteps, which are the four adjacent frames spanning the interval $[t, t+3]$. The GS Decoder processes frames [t+3, t+6] as context in next iterations and continues this progressive reconstruction strategy until reaching the end of the sequence.
To assess its zero-shot depth estimation, we employ two metrics: Absolute Relative Error (AbsRel) and $\delta_1$, where $\delta_1$ represents the percentage of pixels satisfying $\max (\frac{d}{\hat{d}}, \frac{\hat{d}}{d}) < 1.25$, shown in Table \ref{tab:comp:storm:gen}. 
Since ground truth depth is unavailable for generated results, we utilize Depth Anything V2 \cite{yang2024depth} to produce pseudo ground-truth depth maps. While these metrics provide valuable comparative insights, we acknowledge their limitation in assessing absolute depth accuracy, which remains an open challenge in generative depth evaluation. We provide more qualitative results in Figure~\ref{fig:vis_depth_app_0},\ref{fig:vis_depth_app_1}.

\subsubsection{More Qualitative Results}
\begin{figure}[t]
    \centering \subfloat{\includegraphics[width = 0.99\linewidth]{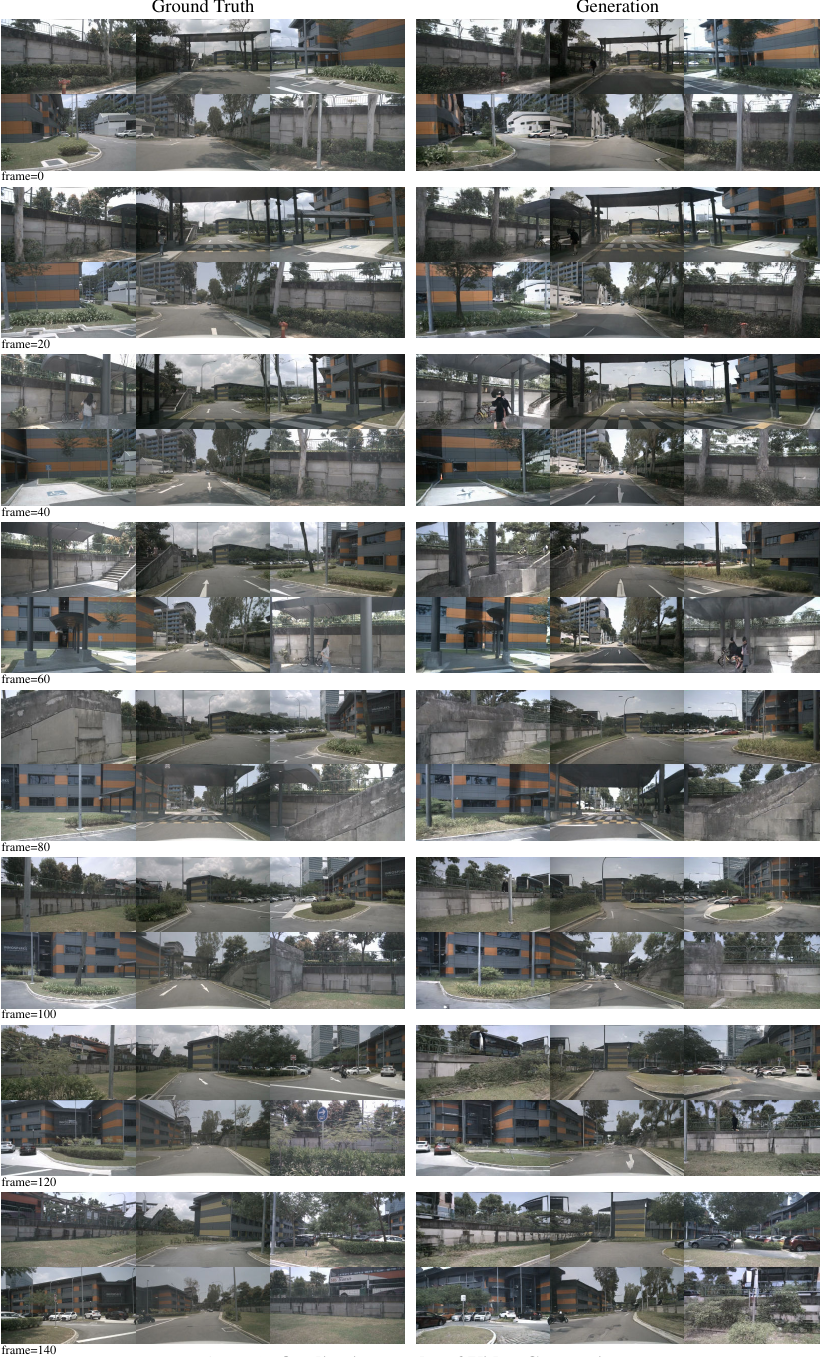}} 
    \vspace{-0.4cm}
    \caption{\textbf{Qualitative results of Video Generation}}
    \label{fig:vis_norefgen_app_0}
\end{figure}

\begin{figure}[t]
    \centering \subfloat{\includegraphics[width = 0.99\linewidth]{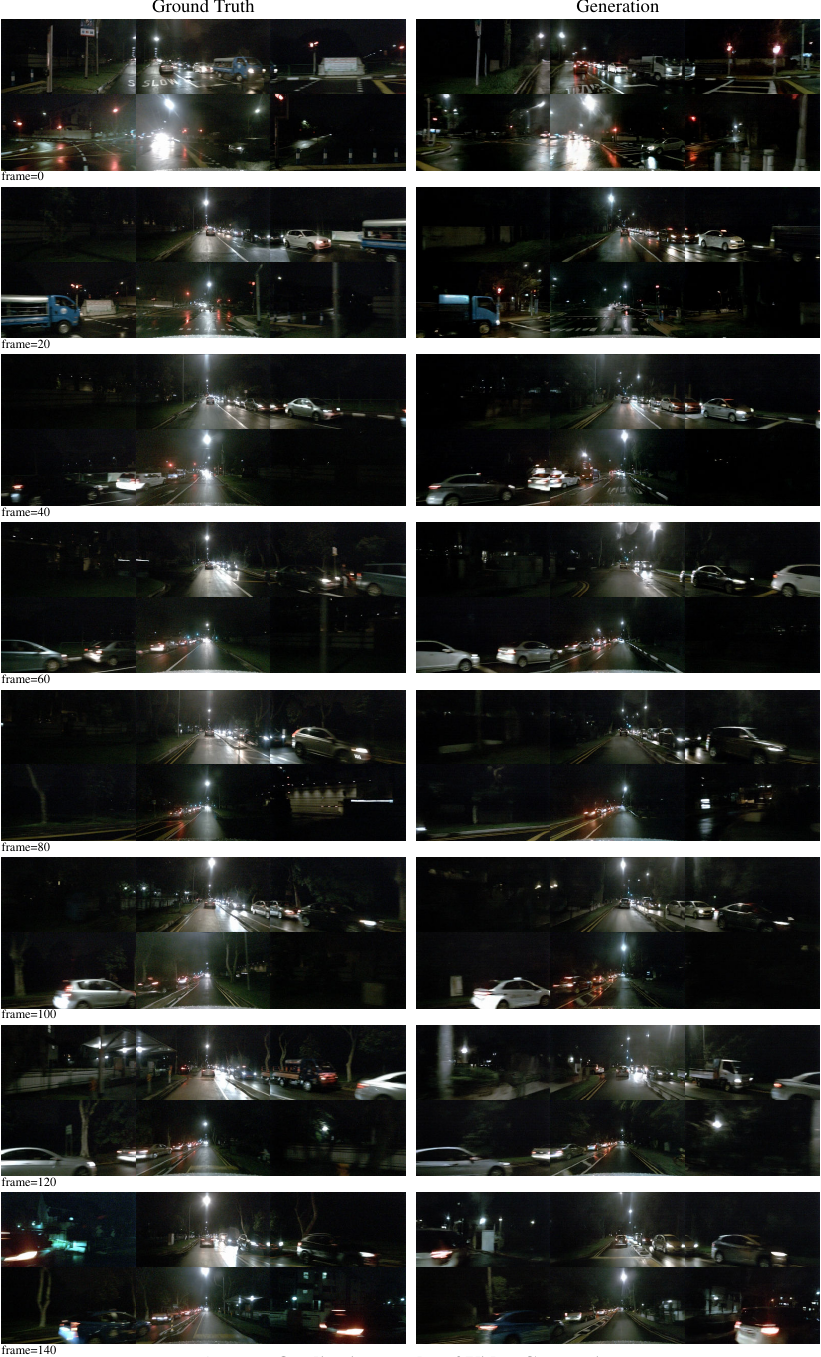}} 
    \vspace{-0.4cm}
    \caption{\textbf{Qualitative results of Video Generation}}
    \label{fig:vis_norefgen_app_1}
\end{figure}

\begin{figure}[t]
    \centering \subfloat{\includegraphics[width = 0.99\linewidth]{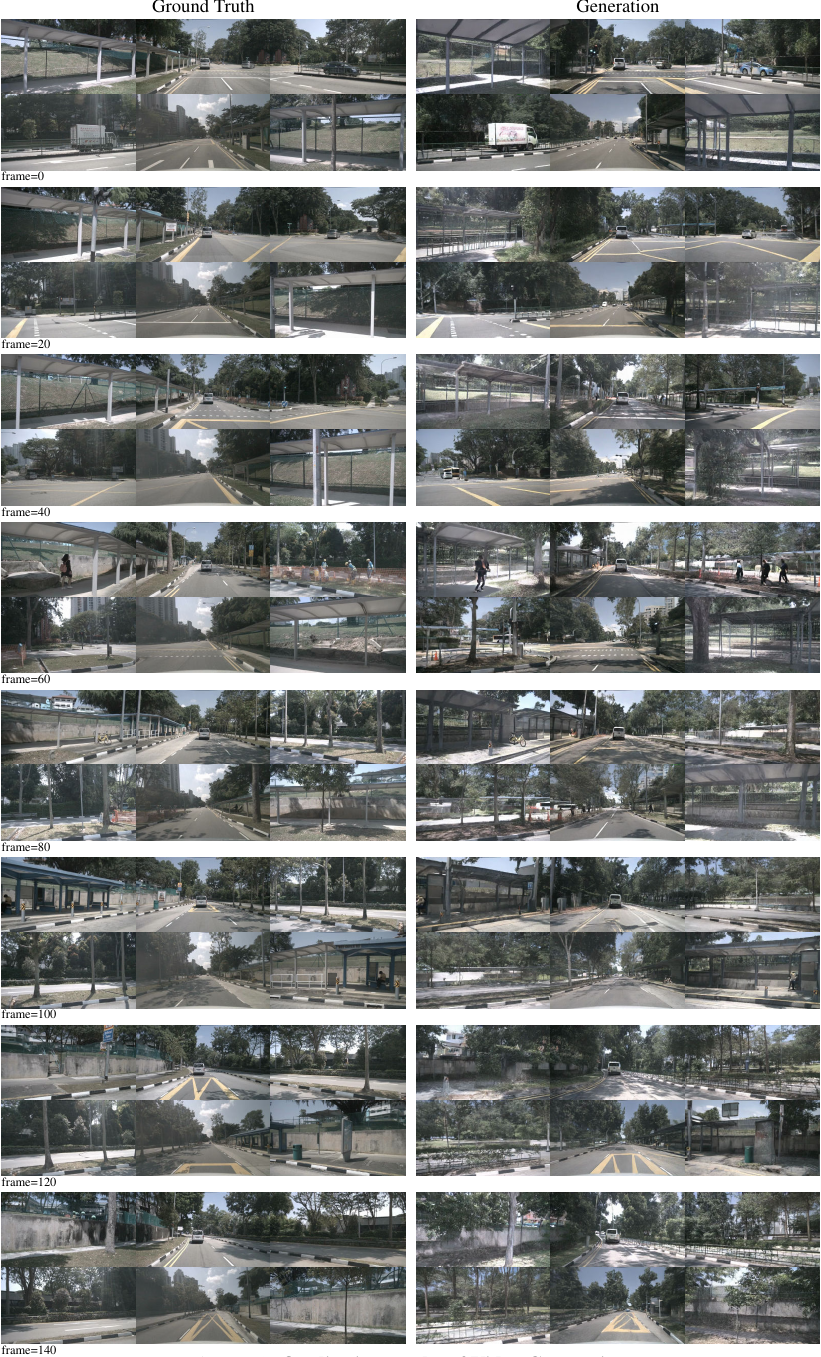}} 
    \vspace{-0.4cm}
    \caption{\textbf{Qualitative results of Video Generation}}
    \label{fig:vis_norefgen_app_2}
\end{figure}

\begin{figure}[htbp]
  \centering
  \includegraphics[width=0.9\textwidth, page=1]{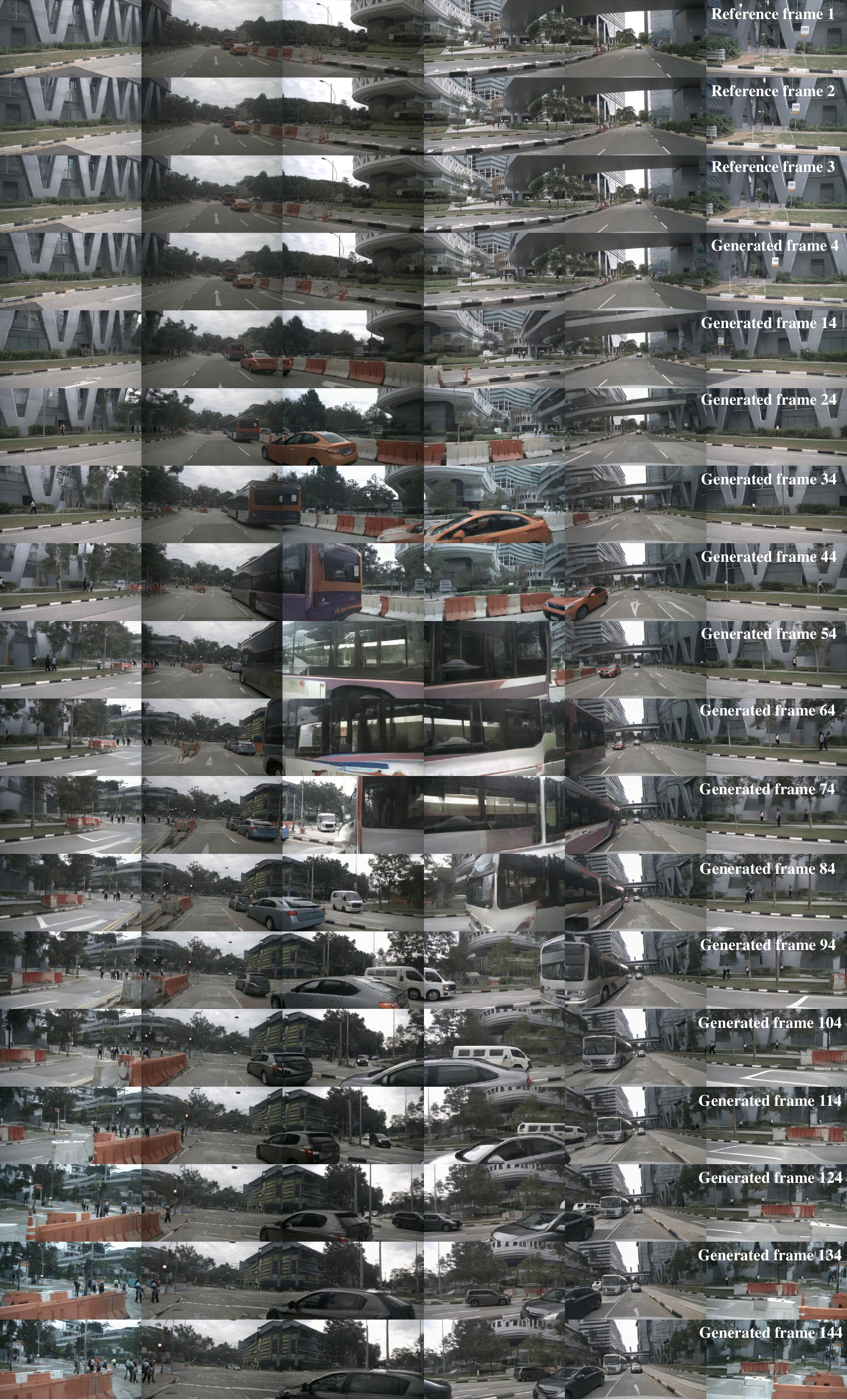}
  \caption{\textbf{Qualitative results of Video Generation from 3 reference frames.}}
  \label{fig:refer_long}
\end{figure}

\begin{figure}[htbp]
  \centering
  \includegraphics[width=0.9\textwidth, page=1]{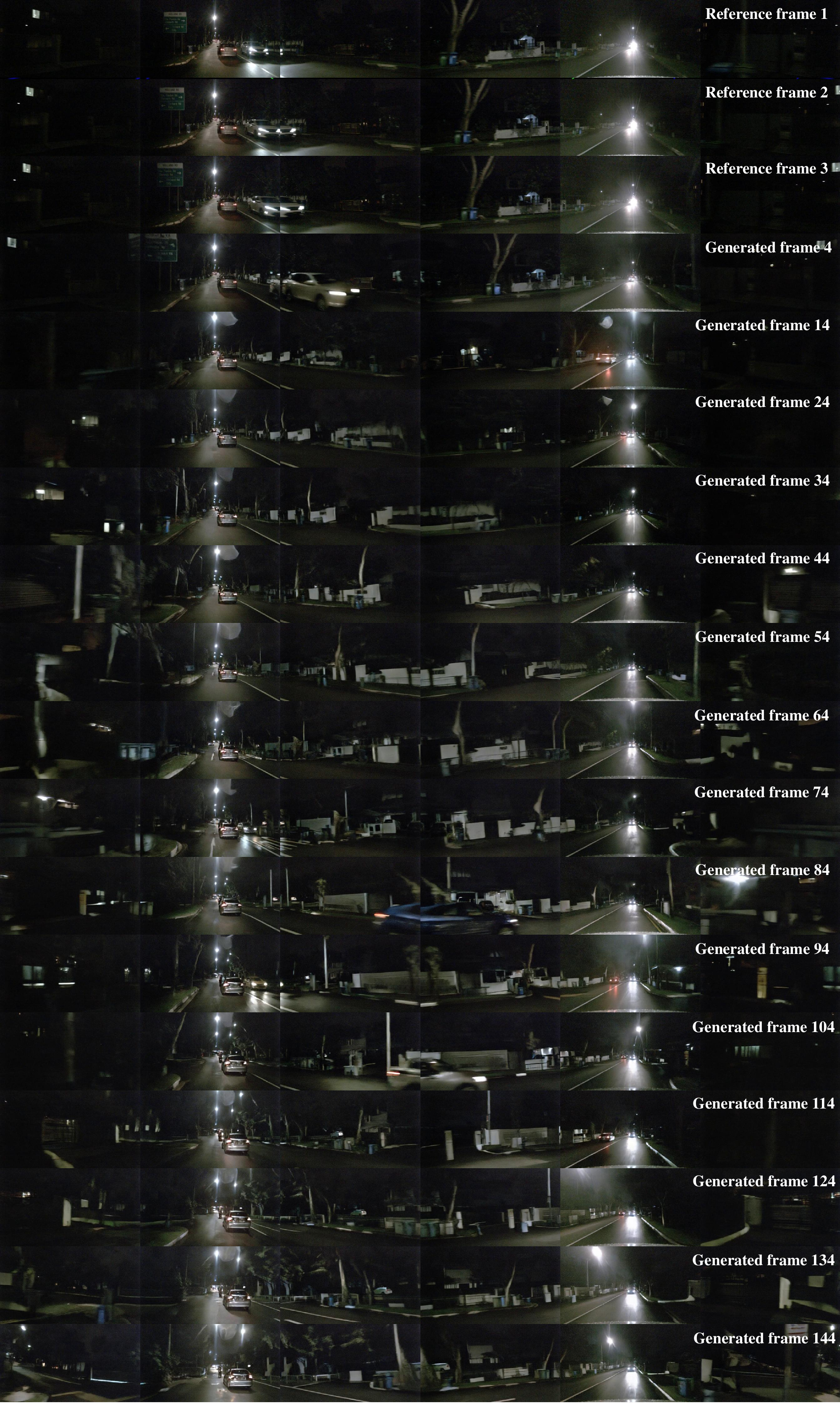}
  \caption{\textbf{Qualitative results of Video Generation from 3 reference frames at night.} Our model imitated the blur of fast motion.}
  \label{fig:night_refer_long}
\end{figure}

We provide more qualitative results in Figure~\ref{fig:vis_norefgen_app_0},~\ref{fig:vis_norefgen_app_1},~\ref{fig:vis_norefgen_app_2},~\ref{fig:refer_long},~\ref{fig:night_refer_long}.

\vspace{-0.5cm}

\vspace{-0.5cm}

\vspace{-1.5cm}

\vspace{-1.5cm}

\vspace{-1.5cm}

\end{document}